\documentclass{article}

\usepackage{microtype}
\usepackage{graphicx}
\usepackage{subcaption}
\usepackage{booktabs} 

\usepackage{hyperref}

\usepackage[preprint]{icml2026}

\usepackage{amsmath}
\usepackage{amssymb}
\usepackage{mathtools}
\usepackage{amsthm}

\usepackage[capitalize,noabbrev]{cleveref}

\theoremstyle{plain}

\theoremstyle{definition}

\theoremstyle{remark}

\usepackage[textsize=tiny]{todonotes}

\usepackage{algorithm}
\usepackage{pifont}
\usepackage{multirow}
\usepackage{wrapfig}
\usepackage[table]{xcolor}
\definecolor{aquamarine}{rgb}{0.82,0.98,0.85}
\definecolor{my_c1}{HTML}{e3f0a3}
\definecolor{my_c2}{HTML}{d0f4e7}
\newcommand{\RETURN}{\STATE \textbf{return} }

\icmltitlerunning{Submission and Formatting Instructions for ICML 2026}

\begin{document}

\twocolumn[
  \icmltitle{\textit{RAP}: \underline{R}untime \underline{A}daptive \underline{P}runing for LLM Inference}



  \icmlsetsymbol{equal}{*}

  \begin{icmlauthorlist}
    \icmlauthor{Huanrong Liu}{equal,um}
    \icmlauthor{Chunlin Tian}{equal,um}
    \icmlauthor{Xuyang Wei}{uestc}
    \icmlauthor{Qingbiao Li}{um}
    \icmlauthor{Li Li}{um}
  \end{icmlauthorlist}

  \icmlaffiliation{um}{Faculty of Science and Technology, University of Macau, Macau, China}
  \icmlaffiliation{uestc}{School of Information and Software Engineering, University of Electronic Science and Technology of China, Chengdu, China}

  \icmlcorrespondingauthor{Qingbiao Li}{qingbiaoli@um.edu.mo}
  \icmlcorrespondingauthor{Li Li}{llili@um.edu.mo}

  \icmlkeywords{Efficient LLMs, ICML}

  \vskip 0.3in
]



\printAffiliationsAndNotice{}  

\newcommand{\model}[1]{\textit{RAP}}
\begin{abstract}
Large language models (LLMs) excel at language understanding and generation, but their enormous computational and memory requirements hinder deployment. 
Compression offers a potential solution to mitigate these constraints. However, most existing methods rely on fixed heuristics and thus fail to adapt to runtime memory variations or heterogeneous KV cache demands arising from diverse user requests.
To address these limitations, we propose \model~, an elastic pruning framework driven by reinforcement learning (RL) that dynamically adjusts compression strategies in a runtime-aware manner. 
Specifically, \model~ dynamically tracks the evolving ratio between model parameters and KV cache across practical execution. Recognizing that FFNs house most parameters, whereas parameter‑light attention layers dominate KV cache formation, the RL agent retains only those components that maximize utility within the current memory budget, conditioned on instantaneous workload and device state.
Extensive experiments results demonstrate that \model~ outperforms state-of-the-art baselines, marking the first time to jointly consider model weights and KV cache on the fly.

\end{abstract}
\section{Introduction}
\label{section1}
Large language models (LLMs) has revolutionized artificial intelligence through unprecedented performance in complex language tasks~\citep{brown2020language, achiam2023gpt, microsoft_bing_new_features,github_copilot}. The autoregressive architectures, however, pair ``billion-parameter'' with memory‑intensive Key–Value (KV) cache, inflating both computation and memory footprints~\citep{fedus2022review,patterson2021carbon,llama_2,PaLM,gemma}. 
While cloud solutions mitigate some burdens, emerging edge scenarios,  mobile devices and real‑time services~\citep{yuan2023mobile,lin2022device,lin2024awq}, demand on‑device inference that current LLMs cannot sustain. Model compression is widely used to preserve generative quality while slashing resource costs.

\begin{figure}[!t]
    \centering
    \includegraphics[width=0.95\linewidth]{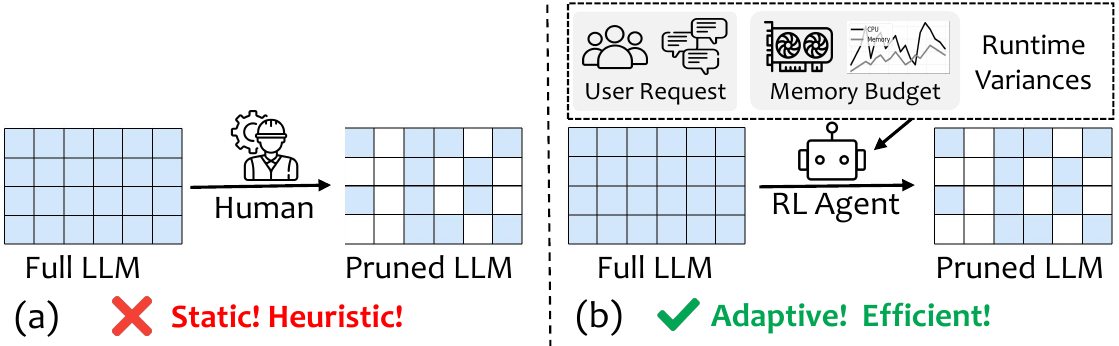}
    \caption{Illustration of \model~. (a) Conventional pruning relies on hand-developed heuristics that focus solely on model weights. (b) \model~ employs a runtime-adaptive RL agent that dynamically prunes LLMs based on real-time user requests and memory budget constraints.}
    \label{fig:pruning_compare}
    \vspace{-1.5em}
\end{figure}

To address LLM deployment bottlenecks, three main compression families have emerged: model pruning \citep{ma2023llmpruner,zhong2024blockpruner,sun2024simpleeffectivepruningapproach,shao2024one}, knowledge distillation \citep{sun2019patient,xu2024survey,chen2024bge}, and quantization~\citep{liu2024spinquant,lin2024awq}. Among these approaches, we primarily focus on pruning. Existing schemes\citep{ma2023llmpruner,zhong2024blockpruner,sun2024simpleeffectivepruningapproach,shao2024one,ashkboos2024slicegpt,gao2024disp,men2024shortgpt,he2024matters,jaiswal2024ffn}, whether element‑, block‑, or layer‑wise, achieve impressive parameter reductions but assume static workloads and rely on heuristic policies, neglecting runtime variability, as shown in Figure~\ref{fig:pruning_compare}. Such rigidity overlooks two dominant sources of autoregressive inference runtime variance: 1) Input‑driven variance: batch size and sequence length directly scale the KV cache memory (e.g., Llama‑7B \citep{touvron2023llama} requires 32 GB of KV cache memory, $\text{batch}=16$ and $\text{length}=4\text{k}$ tokens, dwarfing the static 14 GB model parameters. 2) System‑level variance. Edge devices often exhibit stochastic runtime variance, for instance, interference from co-running applications, affecting available memory budgets on the fly. This situation presents a compelling research question:

\textit{How to select optimal LLM pruning policy that can adapt to heterogeneous, time‑varying request workloads while satisfying fluctuating memory budgets?}

In this paper, we propose \model~, a runtime-adaptive pruning framework that addresses these challenges. \model~ abandons static one-size-fits-all compression in favor of dynamically adjusting the model's sparsity level for each inference. As shown in Figure \ref{fig:pruning_compare}, it introduces a reinforcement learning (RL) agent that observes real-time signals, such as input sequence length, batch characteristics, and current memory availability, and selects an appropriate pruning policy on the fly. This design ensures that the model stays within memory budgets under tight conditions while preserving as many parameters as possible when resources allow. By coupling compression decisions with the execution context, \model~ effectively accommodates heterogeneous workloads and fluctuating system constraints that are impractical for fixed pruning strategies. We formulate adaptive pruning as a sequential decision process and train the RL agent to maximize efficiency without compromising output quality. The agent's reward function balances memory savings against generation fidelity, encouraging policies that reduce memory usage only to the extent they do not degrade performance. Once trained, the agent serves as an intelligent controller during inference, guiding the LLM to prune different components (e.g., attention modules, feed-forward networks, or even entire layers) in response to each request's needs. Notably, \model~ adds negligible runtime cost, since the learned policy can rapidly compute pruning decisions. This yields a flexible, context-aware compression mechanism that seamlessly scales LLM deployments to edge environments. Our experiments demonstrate that \model~ outperforms static pruning baselines across a range of deployment scenarios. Without manual retuning, \model~ adapts to varying batch sizes and sequence lengths, consistently meeting fluctuating memory limits while maintaining strong task performance. For example, under stringent memory constraints, \model~ prunes a substantial fraction of the model’s weights to fit an LLM on-device yet maintains accuracy comparable to an unpruned model. Conversely, when memory is abundant, \model~ leaves the model largely intact to maximize accuracy, effectively achieving the best of both worlds. In summary, our contributions are as follows:

$\bullet$ We propose \model~, a novel runtime-adaptive LLM pruning framework that dynamically adjusts model size based on real-time input demands and memory constraints.

$\bullet$ We cast the pruning policy selection as a reinforcement learning problem and develop an RL agent that learns an optimal policy balancing memory efficiency and model fidelity.
    
$\bullet$ We demonstrate through extensive experiments that \model~ consistently outperforms static compression strategies under dynamic workloads, achieving superior memory savings and faster inference with minimal impact on output quality.

\section{Background and Related Work}
\label{section2}

\subsection{Runtime LLM Inference Memory Breakdown}
\label{section2.1}
Transformer‑based LLMs comprise a stack of \textit{homogeneous} decoder layer, each with a multi‑head attention (MHA) block followed by a feed‑forward network (FFN) block. 

Given that FFNs typically contain approximately $2\times$ the parameters of their corresponding attention modules, the static parameter memory allocation is predominantly determined by FFN weights, which remain fixed once model are loaded. During inference, each token $x$ is projected with $W_{q}$, $W_{k}$, and $W_{v}$ within MHA to obtain
$Q = xW_{q}$, $K = xW_{k}$, and $V = xW_{v}$; the resulting $K$ and $V$ tensors are appended to the KV cache across all layers. For Llama2‑7B ($n_{\text{layers}}=32$, $n_{\text{heads}}=32$, $d_{\text{head}}=128$), the per‑token cache cost is
\(\text{Memory}_{\text{KV}}
= 2\, n_{\text{layers}}\, n_{\text{heads}}\, d_{\text{head}}\, p_{a}
\;\approx\; 0.5~\text{MB}, \)
where the factor~2 stores both keys and values.

Figure~\ref{fig:parameters_kvcache} shows memory footprint across batch size and sequence length. Each pie chart illustrates the relative proportion of memory consumed by model parameters (FFN in orange, MHA in blue) and KV cache (gray). As batch size and sequence length gradually extend, memory consumption transitions from parameter-dominated regimes to KV cache-dominated, highlighting the dynamic nature of memory bottlenecks in practical deployment. Once model is loaded into memory, increasing the batch size or extending the context length does not affect parameter memory consumption but substantially increases KV cache memory overhead.


\begin{equation}
\text{KV cache} \propto (\text{batch size}) \times (\text{sequence length}) \times n_{\text{layers}}
\end{equation}

Therefore, practical memory scaling is driven almost entirely by the MHA-generated KV cache, underscoring the need for adaptive compression schemes that address both the FFN‑heavy static parameter and this rapidly expanding dynamic cache.


\subsection{Existing LLM Pruning}

For runtime LLM inference, pruning strategies~\citep{hu2021lora, liu2023deja, xia2023flash, yin2023outlier, zhang2023loraprune} must balance efficiency, accuracy, and adaptability.
\textbf{1) Static vs. dynamic pruning:} Static methods (e.g., ISC~\citep{das2023beyond}, SparseGPT~\citep{frantar2023sparsegpt}, E-Sparse~\citep{li2023sparse}, Wanda~\citep{sun2024simpleeffectivepruningapproach}) apply fixed sparsity without retraining, achieving up to 50\% sparsity but degrading under higher sparsity levels and fundamentally lacking adaptability. Structured variants~\citep{ashkboos2024slicegpt, chen2023lorashear, ma2023llmpruner, zhao2024apt} improve hardware efficiency but require retraining (e.g., LoRA~\citep{hu2021lora}). In contrast, dynamic pruning~\citep{an2024fluctuation, federici2024efficient, le2025probe, liu2023deja} adapts per input, improving flexibility but retaining full weights and inducing irregular sparsity, limiting hardware speedups.
\textbf{2) Parameter-only vs. parameter+KV compression:} Most pruning reduces weights~\citep{ma2023llmpruner, ashkboos2024slicegpt, li2023sparse, sun2024simpleeffectivepruningapproach} but ignores KV cache, a major runtime bottleneck. While weight pruning shrinks static parameter, it fails under long-context due to exponentially growing KV cache. Recent methods (e.g., ShortGPT~\citep{men2024shortgpt}, BlockPruner~\citep{zhong2024blockpruner}, LLM-Drop~\citep{he2024matters}, FinerCut~\citep{zhang2024finercut}) prune both parameters and KV cache, reducing computation and memory but often rely on static rules, sacrificing accuracy. The core trade-off persists: parameter-only pruning is insufficient, while aggressive KV cache pruning hurts performance.
\textbf{3) Heuristic vs. learning-based control:} Heuristic methods~\citep{sun2024simpleeffectivepruningapproach, frantar2023sparsegpt, ma2023llmpruner} use static scores (e.g., magnitude, saliency), lacking runtime adaptability or end-to-end optimization. Learning-based policies can jointly optimize for speed, memory, and accuracy. Though RL has proven effective in~\citep{andrychowicz2020learning, mnih2015human, zhang2017deep}, it remains underexplored for LLM pruning, particularly for coordinated control of parameter and KV cache. \model~ addresses this by introducing an RL-based policy that dynamically prunes both components, enabling runtime-adaptive and efficient inference beyond static baselines.
\section{Motivation}

\label{section3}
In this section, we present key observations from stochastic workloads, model-intrinsic and system factors for practical LLM inference.

\begin{figure}[!t]
    \centering
    \includegraphics[width=0.49\linewidth]{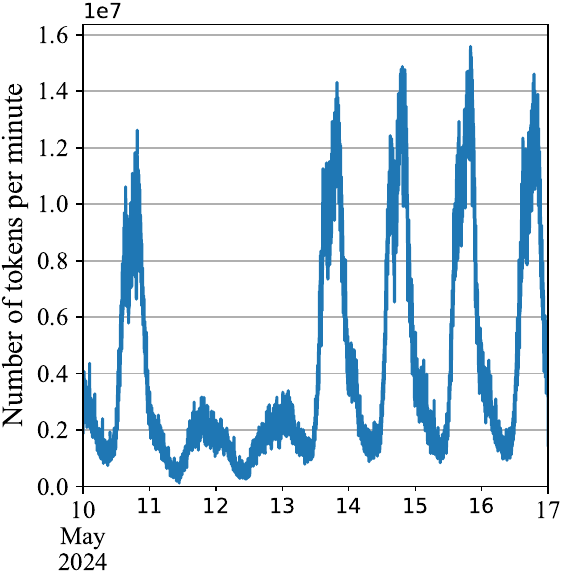}
    \includegraphics[width=0.49\linewidth]{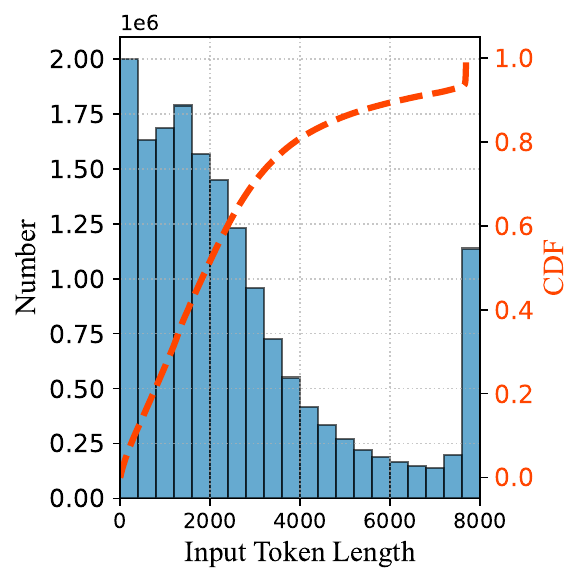}
    \caption{Distribution and daily variation of a conversational
    LLM inference workload.}
    \label{fig:token_dis}
\end{figure}

\begin{figure}[!t]
    \centering
    \includegraphics[width=0.95\linewidth]{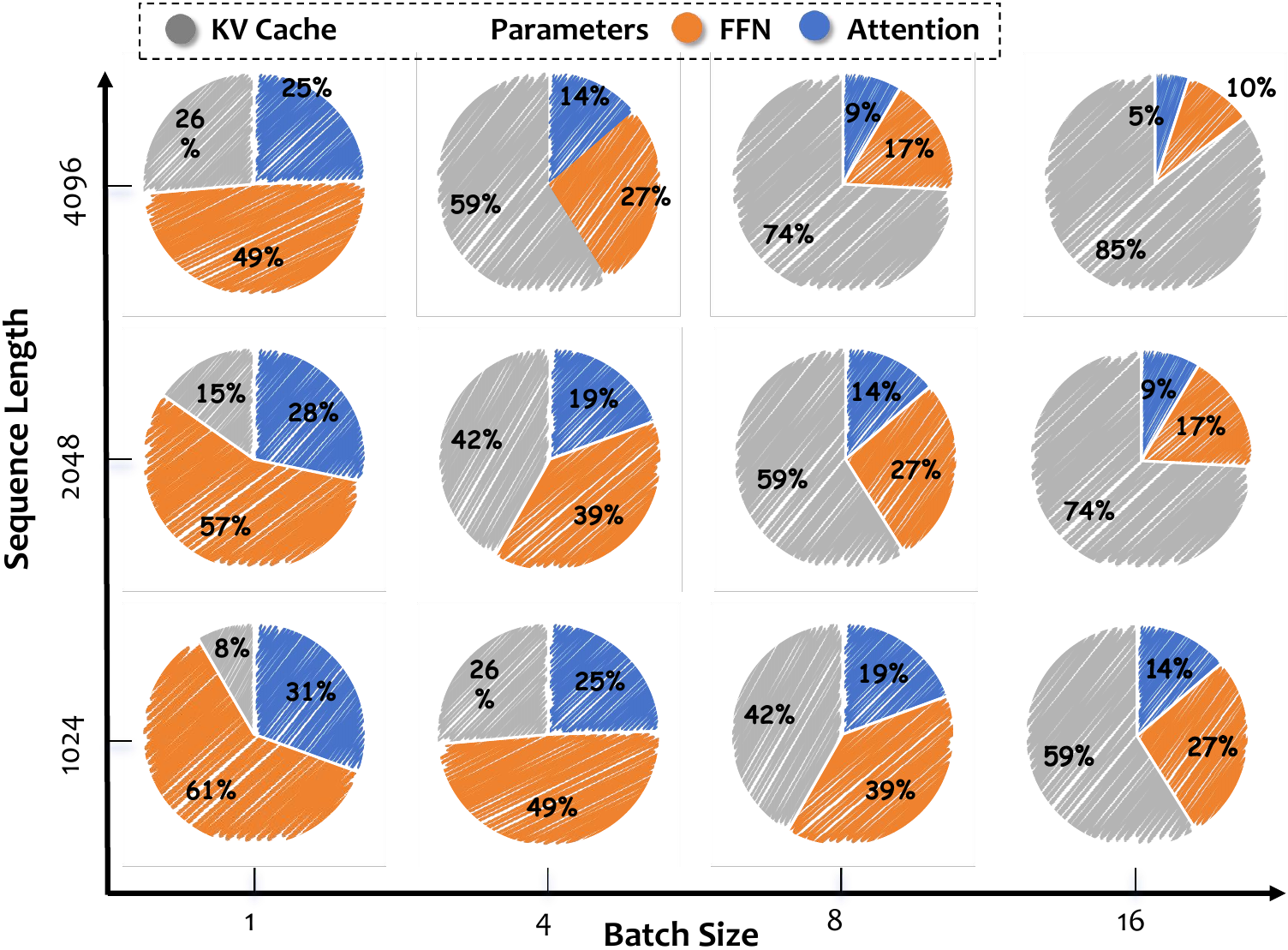}
    \caption{Dynamic memory footprint across varying batch sizes and sequence lengths.}
    \label{fig:parameters_kvcache}
\end{figure}

\noindent $\blacktriangleright$ Takeaway 1: \textit{Runtime workloads are inherently dynamic.}
Modern LLM service platforms must operate under highly volatile workload conditions \citep{patel2024splitwise,li2024llm,jaiswal2025serving}. Figure \ref{fig:token_dis}, derived from an Azure LLM‑inference trace \citep{trace}, reveals that prompt‑length distributions and request arrival rates fluctuate markedly over time, producing a non‑deterministic mix of short conversational turns and burst, long‑form inputs. Figure \ref{fig:parameters_kvcache} illustrates how memory allocation transitions from parameter-dominated regimes at low batch sizes to KV cache-dominated scenarios as batch size and sequence length scale up, fundamentally reshaping inference memory bottlenecks. These findings reveal a fundamental limitation in current serving infrastructures: static resource allocation and heuristic per-request throttling mechanisms fail to satisfy Quality of Experience (QoE) demands for latency and memory efficiency when facing the inherently dynamic and unpredictable nature of real-time inference workloads.

\begin{figure}[!t]
    \centering
    \includegraphics[width=1\linewidth]{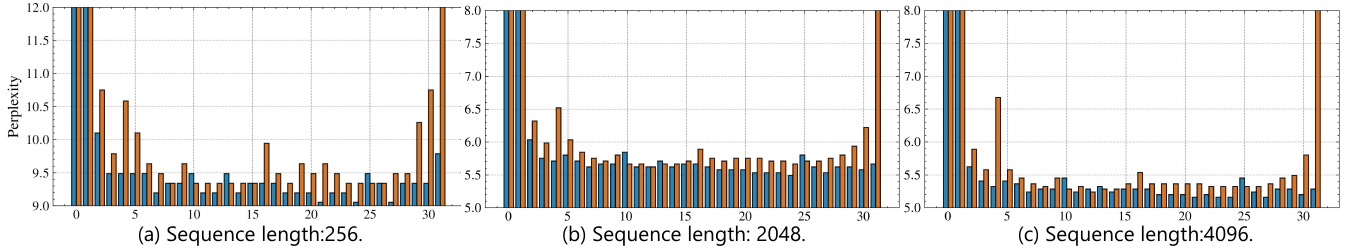}
    \captionof{figure}{Block sensitivity analysis: removing specific MHA and FFN under diff. sequence length.}
    \label{fig:ppl}
\end{figure}

\noindent $\blacktriangleright$ Takeaway 2: \textit{Homogeneous blocks exhibit heterogeneous impact.}
Current transformer architectures exhibit seemingly homogeneous layers ($\S \ref{section2.1}$), yet their internal blocks (MHA and FFN) contribute heterogeneously to generation quality. Prior studies have broadly differentiated layer importance \citep{ma2023llm,zhang2024investigating,yao2024layer,pan2024lisa} or assumed fixed superiority of FFN over MHA \citep{he2024matters}. However, as summarized in Figure \ref{fig:ppl}, the impact of MHA and FFN removal on perplexity (PPL) varies significantly across layers, challenging coarse-grained or uniform assumptions. Moreover, solely optimizing FFN cannot alleviate the KV cache bottleneck that arises in long sequences and large batch sizes. Additionally, block importance demonstrates dynamic shifts across different request lengths, underscoring the limitations of existing static, heuristic-based pruning strategies \citep{yao2024layer,pan2024lisa,ma2023llm,men2024shortgpt}. These insights highlight the critical imperative for adaptive method that dynamically discerns and leverages block-level sensitivity to accommodate heterogeneous runtime computational demands.

\begin{figure}[!t]
    \centering
    \includegraphics[width= 0.75\linewidth]{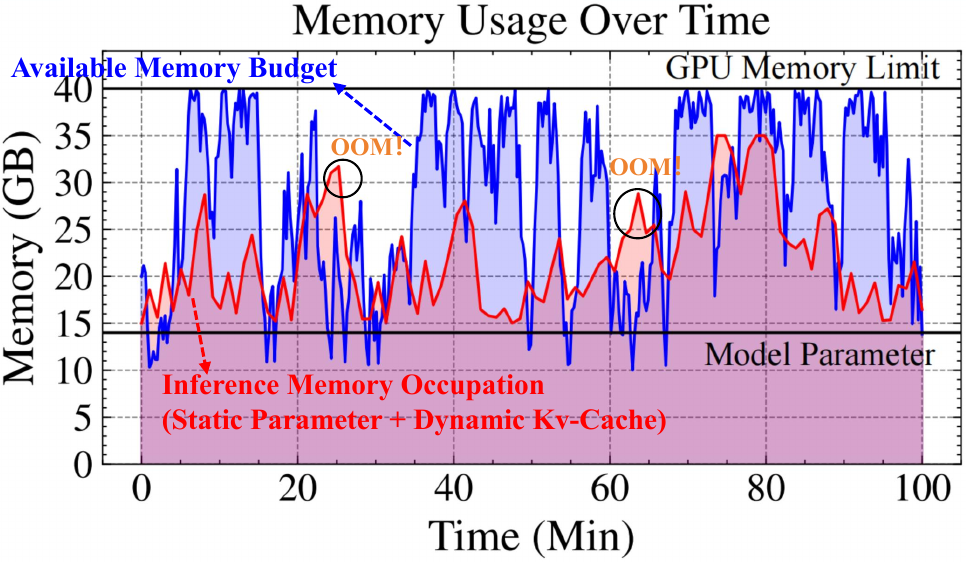}
    \caption{Dynamic memory allocation trace for Llama2-7B on an NVIDIA A40 (40 GB) \citep{nvidia2020a40}. Blue indicates available memory; red shows real-time usage (model + KV cache), which scales with workload and cause out-of-memory (OOM) errors under heavy requests.} 
    \label{fig:memory}
\end{figure}

\noindent $\blacktriangleright$ Takeaway 3: \textit{Real‑world systems demonstrates runtime variance.}
Real-world LLM inference systems rarely maintain consistent memory availability \citep{wang2024burstgpt,yu2023faaswap,xu2022igniter}. Instead, they encounter dynamic memory variability driven by heterogeneous LLM workloads and interference by co-running applications. 
Azure LLM service traces \citep{trace} show that prompt-length and request-arrival spikes can induce instantaneous GPU-memory fluctuations of up to 5–10. Concurrent workloads further disrupt cache and bandwidth allocation, exacerbating contention and latency instability. As Figure~\ref{fig:memory} illustrates, these memory surges often preempt co-running applications and invalidate the fixed-budget assumptions of existing pruning and scheduling methods, highlighting a critical gap between current serving frameworks and real-world, memory-dynamic inference environments.  


\begin{figure}[!t]
    \centering
    \includegraphics[width=0.95\linewidth]{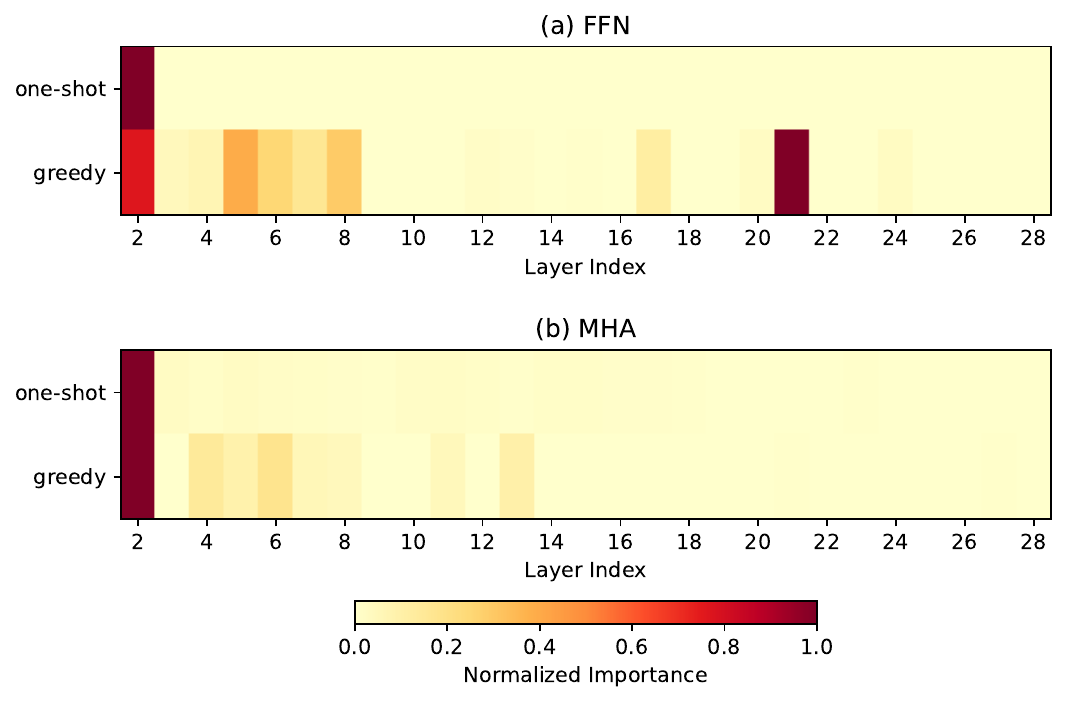}
    \caption{Per‑block perplexity sensitivity (FFN vs. MHA) under one‑shot and GSI pruning, with GSI revealing inter‑layer heterogeneity missed by static one‑shot methods.}
    \label{fig:heatmap}
\end{figure}
\section{\model~ Design}

\label{section4}
In this section, we present the design of \model~. Specifically, we first introduce greedy sequential importance analysis $\S \ref{section4.1}$ to thoroughly assess the impact of individual transformer blocks. Then, we explain how we formulate the problem of runtime dynamic pruning as an RL task $\S \ref{section4.2}$.

\subsection{Greedy Sequential Importance}
\label{section4.1}

As discussed in $\S \ref{section3}$, blocks exhibit heterogeneous contributions to model performance. Conventional one-shot pruning methods \citep{ma2023llmpruner, zhong2024blockpruner} that remove layers solely based on individual sensitivity ignore inter-layer dependencies, often leading to substantial performance degrades under aggressive compression ratio. The deep composition of nonlinear activations and residual connections in LLMs induces strong inter-layer dependencies \citep{ling2024slimgpt, meng2024osscar}, rendering the network fragile to architectural changes. Consequently, excising even a single layer can trigger a cascade of representational errors that corrupts the entire model's functionality. To mitigate this, we propose \textit{Greedy Sequential Importance} (GSI) analysis. As detailed in Algorithm~\ref{alg:GSI}, GSI performs iterative pruning by progressively removing the block whose exclusion results in the minimal deterioration, followed by re-evaluating after each step. This step-wise recalibration controls error accumulation, stabilizes accuracy over successive pruning stages, and achieves a more balanced compression–performance trade-off compared to static, one-shot methods. Figure~\ref{fig:heatmap} further highlights that one-shot pruning neglects inter-layer heterogeneity, leading to suboptimal pruning decisions. In this paper, we select perplexity as the proxy metric for the GSI algorithm to measure the impact of block removal on overall model performance, since perplexity is a widely-accepted metric for generative capabilities of LLM. Alternatively, task-specific downstream metrics can serve as a proxy to enable pruning decisions more aligned with target scenarios. Overall, GSI offers a principled and adaptive approach to LLM compression, effectively balancing model size reduction with task performance preservation.





\begin{figure*}[!t]
    \centering
    \includegraphics[width= 0.75\linewidth]{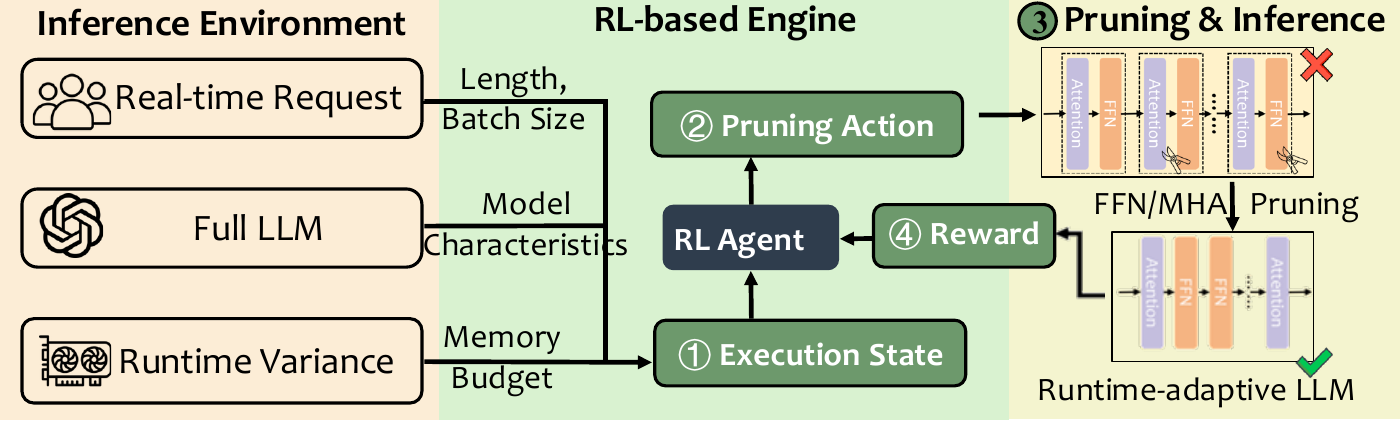}
    \caption{Design overview of \model~. (1) Runtime statistics from inference environment are encoded into execution state. (2) RL agent selects FFN/MHA blocks for pruning. (3) Resulting memory consumption and performance constitute the reward. (4) Agent gains reward and reinforces, completing an online loop for dynamically balanced efficiency and accuracy.}
    \label{fig:framework}
\end{figure*}

\subsection{RL-Guided Pruning Decisions}

\label{section4.2}
To address the dynamic inference environments characterized by user-request workloads and system runtime variance, we propose \model, an adaptive pruning framework based on reinforcement learning. Figure~\ref{fig:framework} presents the design overview of \model~. 
\ding{172}~At each inference step, \model~ observes the real-time request characteristic, model configuration, and available memory budget to determine the current execution state. 
\ding{173}~Based on this state, the RL agent selects a pruning policy that satisfies the memory constraint while aiming to preserve model performance. 
\ding{174}~The base model then executes the selected pruning policy by removing the corresponding FFN and MHA blocks, and subsequently performs inference on the compressed architecture.
\ding{175}~Finally, \model~ evaluates inference metrics, including memory overhead and perplexity, to derive a reward that quantifies how effectively the selected action balances computational efficiency with model performance. We next define the core RL components, \textit{State}, \textit{Action}, and \textit{Reward}, to formalize the optimization space of \model~.

\textsc{\textbf{State:}} the state at the $t$-th timestep $S_{t} = (s_{t}^{Req},s_{t}^{Model},s_{t}^{Sys}) \in \mathbf{S}$ consists of three components:

\begin{itemize}
    \item $s_{t}^{Req} = (R_{bs}, R_{sql})$ captures the real-time request characteristics, consisting of the batch size $R_{bs}$ and sequence length $R_{sql}$.

    \item $s_{t}^{Model} = \left( \{\text{MHA}_{imp,i}\}_{i=1}^{N}, \{\text{FFN}_{imp,i}\}_{i=1}^{N} \right)$, representing importance score of each MHA and FFN block computed via GSI algorithm $\S \ref{section4.1}$, where $N$ denotes the total number of blocks, capturing the granular block-level model configuration.

    \item $s_{t}^{Sys} = (Sys_{avail}, Sys_{req})$ represents the runtime system memory state, where $Sys_{avail}$ denotes the currently available system memory, and $Sys_{req}$ indicates the anticipated memory overhead after executing the selected pruning policy.

\end{itemize}

\textsc{\textbf{Action:}} 
At each pruning step $t$, given $N$ layers model and input state $S_t$, the action set is defined as $A_t = \{(a_1, \dots, a_{2N}) \mid a_i \in \{0,1\}\}$, where $2N$ binary indicators represent whether to retain ($a_i = 1$) or remove ($a_i = 0$) each of the $2N$ transformer blocks (one MHA and one FFN per layer) at step $t$. Simultaneous multi-block selection creates an intractable action space of size $2^{2N}$; for example, Llama2-7B with 64 blocks, this yields approximately $2^{64} \approx 1.8 \times 10^{19}$ possible actions. To address this computational challenge, we decompose the decision into sequential single-block selections, reducing the action space to $2N$ decision step. However, directly applying one-shot top-$k$ pruning proves suboptimal, as demonstrated in $\S ~\ref{section4.1}$, since block importance dynamically changes after each removal. More precisely, we utilize GSI-derived importance scores to iteratively remove the least important block at each step. After each removal, we re-assess the importance hierarchy among remaining blocks and select the next least important candidate, repeating this greedy selection until the peak memory footprint meets the memory budget constraint. While the approach requires iterative decision-making, the RL agent employs a lightweight 2-layer MLP with minimal parameters, ensuring computational overhead remains negligible compared to the inference costs of billion-parameter LLMs.

\begin{algorithm}[t]
\caption{Greedy Sequential Importance (GSI) using perplexity as the proxy metric}
\label{alg:GSI}
\begin{algorithmic}[1]
\REQUIRE Pre-trained model $\mathcal{M}$, evaluation corpus $\mathcal{C}$, target prune ratio $\rho$
\ENSURE Pruned model $\mathcal{M}^{(t)}$, pruned blocks $\{B_{k_t}\}$, perplexities $\{P_{k_t}\}$
\STATE $\mathcal{M}^{(0)} \gets \mathcal{M}$,\quad $t \gets 0$ 
\WHILE{$\mathrm{PruneRatio}(\mathcal{M}^{(t)}) < \rho$}
  \STATE $t \gets t + 1$
  \FORALL{block $B_i$ in $\mathcal{M}^{(t-1)}$}
    \STATE $\hat{\mathcal{M}}_{i} \gets \mathcal{M}^{(t-1)} \setminus B_i$ 
    \STATE $P_i \gets \exp\!\Bigl(-\tfrac{1}{|\mathcal{C}|} \sum_{w \in \mathcal{C}} \log p_{\hat{\mathcal{M}}_{i}}(w)\Bigr)$ 
  \ENDFOR
  \STATE $k \gets \arg\min_{i} P_i$ 
  \STATE $\mathcal{M}^{(t)} \gets \hat{\mathcal{M}}_{k}$
\ENDWHILE
\RETURN $\mathcal{M}^{(t)},\;\{B_{k_1},\dots,B_{k_t}\},\;\{P_{k_1},\dots,P_{k_t}\}$
\end{algorithmic}
\end{algorithm}

\textsc{\textbf{Reward:}} 
To ptimize a pruned model under a fixed memory budget presents a multi-objective challenge. We address this by formulating a unified reward as a weighted sum of two specialized metrics: $R_{\mathrm{ppl}}$, reflecting language modeling performance via perplexity, and $R_{\mathrm{mem}}$, which penalizes peak memory consumption during inference.

\begin{equation}
R_t = \sum_{i=1}^{2N} (A_t)_i \left( \alpha R^{\mathrm{ppl}}_{i} - \beta R^{\mathrm{mem}}_{i} \right)
\label{eq:reward}
\end{equation}

Here, $N$ is the total number of blocks. At each time step $t$, $A_t$ is a binary action vector where $(A_t)_i=1$ indicates that block $i$ is preserved while $(A_t)_i=0$ will remove block $i$. The terms $R^{\mathrm{ppl}}_{i}$ (detailed in $\S$~\ref{section4.2}) and $R^{\mathrm{me}}_{i}$ denote the perplexity importance and estimated memory footprint of block $i$, respectively. The hyperparameters $\alpha$ and $\beta$ act as reward scale factors, tuning the accuracy–memory trade-off and penalizing bottleneck workloads when necessary. Specifically, in this paper, we set $\alpha =1$, $\beta =0.3$. A detailed description of RL-agent algorithm can be found in Appendix~\ref{appendix:Algorithm}.

\section{Experiments}
\label{section5}

\subsection{Experimental Setup}
\textbf{Model and Dataset.} We implemented \model~ using PyTorch~\citep{paszke2019pytorch} and the HuggingFace Transformers library~\citep{wolf2019huggingface} for managing models and datasets. All experiments were conducted on NVIDIA A40 GPUs~\citep{nvidia2020a40}. For GSI, we used the Alpaca dataset~\citep{alpaca} to compute perplexity importance. We evaluated \model~ over representative LLMs: Llama2-7B~\citep{touvron2023llama2}, Llama3-8B~\citep{dubey2024llama}, Qwen1.5-7B~\citep{bai2023qwen} and Qwen2.5-7B~\citep{yang2024qwen2}. We assessed model performance using the LM Evaluation Harness~\citep{eval-harness} following Llama’s evaluation protocol to perform zero-shot task classification on common sense reasoning datasets: BoolQ~\citep{clark2019boolqexploringsurprisingdifficulty}, PIQA~\citep{Bisk2020_piqa}, HellaSwag~\citep{zellers2019hellaswag}, WinoGrande~\citep{sakaguchi2019winograndeadversarialwinogradschema}, ARC-easy~\citep{allenai:arc}, ARC-challenge~\citep{allenai:arc}, and OpenbookQA~\citep{OpenBookQA2018}. We tested the model generative ability using WikiText2 \citep{merity2016pointer} and PTB \citep{marcus1993building} dataset. A detailed description of the benchmarks can be found in Appendix \ref{appendix:dataset}.

\begin{table*}[!t]
    \centering
    \caption{Zero‑shot performance of pruned versus dense model under different memory budgets. $^{1}$ 100\% memory budget indicates exceeding peak inference memory usage (parameters + KV cache).}
    \resizebox{0.75\linewidth}{!}{\begin{tabular}{cc|cc|ccccccc|c}
    \toprule [1pt]
     \multirow{2}{*}{\textbf{
     Budget}} & \multirow{2}{*}{\textbf{Schemes}}   & \multicolumn{2}{c|}{\textbf{Perplexity} $\downarrow$} & \multicolumn{8}{c}{\textbf{Commonsense Task (\%)} $\uparrow$} \\ 
         & & \textbf{WikiText2} & \textbf{PTB}& \textbf{BoolQ} &\textbf{PIQA} & \textbf{WinoG.} & \textbf{HellaS.} & \textbf{ARC-e} & \textbf{ARC-c} & \textbf{OBQA} & \textbf{Avg.} \\ \midrule [1pt]
      \rowcolor{my_c1}   \multicolumn{12}{c}{\textbf{Llama2-7B}} \\ \toprule [1pt]
        100\%$^{1}$   & Dense  & 5.47 & 24.09 & 77.74 & 79.11  & 68.97  & 75.98  & 74.62  & 46.25 & 44.20 & 66.70 \\ 
        \hline
        \multirow{6}{*}{80\% } & LLMPruner \citep{ma2023llmpruner}  & 28.42 & 278.05 & 50.63  & 68.82  & 54.14  & 52.66  & 49.33 & 30.20 & 34.59 & 48.63\\
         & SliceGPT \citep{ashkboos2024slicegpt} & 58.33  & 211.33  & 61.99  & 65.29  & 58.08  & 43.43 & 52.27 & 32.08 & 28.99 & 48.88\\ 
        & ShortGPT \citep{men2024shortgpt}   & 79.49 & 171.02  & 62.17  & 60.12  & 60.38  & 43.70 & 41.25 & 30.12 & 35.00 & 47.53\\
        & MHA-Drop \citep{he2024matters}   & 1068.99  & 1579.65  & 46.08  & 54.56  & 50.83  & 29.12 & 28.62 & 27.47 & 25.20 & 37.41\\
        & FFN-Skip \citep{jaiswal2024ffn}   & 28720.72  & 32216.40  & 42.80  & 49.35  & 51.22  & 26.69 & 27.18 & 28.49 & 26.60 & 36.05\\
         & \cellcolor{my_c2}\textbf{\model~} & 
         \cellcolor{my_c2}\textbf{11.80} &
         \cellcolor{my_c2}\textbf{46.56} & 
         \cellcolor{my_c2}\textbf{62.81} & 
         \cellcolor{my_c2}\textbf{73.99} & 
         \cellcolor{my_c2}\textbf{63.38} & 
         \cellcolor{my_c2}\textbf{65.77} &
         \cellcolor{my_c2}\textbf{60.35} &
         \cellcolor{my_c2}\textbf{36.69} & 
         \cellcolor{my_c2}\textbf{36.60} & 
         \cellcolor{my_c2}\textbf{57.08}\\ \hline 
      \multirow{6}{*}{60\% } & LLMPruner  & 96.52  & 711.38  & 55.96  & 61.15  & 50.83  & \textbf{38.09} & 35.06 & 27.13 & 28.40 & 42.38\\ 
        & SliceGPT  & 348.26  & 590.12  & \textbf{61.04 } & 54.56  & 49.49  & 28.99 & 30.68 & 23.46  & 25.40 & 39.09 \\ 
        & ShortGPT    & 964.92 & 2219.93  & 55.57  & 50.98  & 50.51  & 27.83 & 26.39 & 27.47 & 26.60 & 37.91\\
         & MHA-Drop & 6731.72  & 7914.86  & 37.83  & 49.89  & 49.88  & 25.72 & 26.05 & 25.43 & 26.80 & 34.51\\
         & FFN-Skip   & 202008.00  & 160986.48  & 44.83  & 50.64  & 48.60  & 25.86 & 25.72 & 28.92 & 28.19 & 36.13\\
          & \cellcolor{my_c2}\textbf{\model~} &   
          \cellcolor{my_c2}\textbf{84.78} & \cellcolor{my_c2}592.65 & \cellcolor{my_c2}57.16 & \cellcolor{my_c2}\textbf{58.26} & \cellcolor{my_c2}\textbf{53.75} & \cellcolor{my_c2}37.81 & \cellcolor{my_c2}\textbf{37.79} & \cellcolor{my_c2}26.02 &
          \cellcolor{my_c2} \textbf{30.20} & \cellcolor{my_c2}\textbf{43.00}\\
        
        \midrule [1pt]
       \rowcolor{my_c1}  \multicolumn{12}{c}{\textbf{Llama3-8B}} \\ \toprule [1pt]
        100\%  & Dense  & 6.13 & 9.91  & 81.35  & 80.78  & 72.61  & 79.14  & 77.69 & 53.33 & 45.00 & 69.99 \\ 
        \hline
        \multirow{6}{*}{80\% } & LLMPruner  & 48.94  & 99.33  & 62.17  & 65.02  & 51.93  & 42.32  & 41.33  & 25.50 & 29.40 & 45.37\\
         & SliceGPT  & 143.93  & 71.99  & 61.87  & 65.94  & 54.05  & 42.51 & 54.37 & 31.06 & 27.80 & 48.23 \\ 
        & ShortGPT & 37412.61 & 41988.11  & 56.82  & 58.59  & 54.85  & 37.71  & 36.99 & 30.20 & 28.00 & 43.30 \\
        & MHA-Drop  & 529.00  & 737.98  & 37.80  & 53.92  & 50.20  & 26.87 & 30.47 & 23.81 & 27.20 & 35.75 \\
        & FFN-Skip   & 164387.73  & 149698.12  & 55.77  & 51.47  & 50.74  & 25.90 & 25.75 & 25.25 & 28.00 & 37.56 \\
         & \cellcolor{my_c2}\textbf{\model~} & 
         \cellcolor{my_c2}\textbf{12.98} &
         \cellcolor{my_c2}\textbf{27.15} & 
         \cellcolor{my_c2}\textbf{68.84} & 
         \cellcolor{my_c2}\textbf{76.55} & 
         \cellcolor{my_c2}\textbf{66.11} & 
         \cellcolor{my_c2}\textbf{65.55} &
         \cellcolor{my_c2}\textbf{62.12} &
         \cellcolor{my_c2}\textbf{39.51} & 
         \cellcolor{my_c2}\textbf{39.60} & 
         \cellcolor{my_c2}\textbf{59.77}\\ \hline 
      \multirow{6}{*}{60\% } & LLMPruner  & 4009.81  & 1882.99  & 40.48  & 50.05  & \textbf{52.33 } & 26.26  & 26.77  & 25.94 & 27.00 & 35.55\\ 
        & SliceGPT   & 2844.28  & 1084.03  & 40.86  & 53.86  & 48.93  & 27.97 & 32.07 & 23.04 & 25.80 & 36.08 \\ 
        & ShortGPT   & 13284.81  & 13512.55  & 41.44  & 50.98  & 50.03  & 26.74 & 25.46 & 25.34 & 26.80 & 35.26\\
         & MHA-Drop   & 1757.11 & 2102.15  & 37.83  & 51.90  & 50.12  & 26.04  & 27.86 & 22.95 & 25.80 & 34.64\\
         & FFN-Skip   & 624965.25  & 765475.19  & 51.93  & 51.95  & 50.03  & 26.02 & 24.03 & \textbf{26.96} & 28.59 & 37.08\\
          & \cellcolor{my_c2}\textbf{\model~} &   \cellcolor{my_c2}\textbf{246.53} & \cellcolor{my_c2}\textbf{355.47} & \cellcolor{my_c2}\textbf{52.20} & \cellcolor{my_c2}\textbf{56.69} & \cellcolor{my_c2}50.36 & \cellcolor{my_c2}\textbf{31.91} & \cellcolor{my_c2}\textbf{33.54} & \cellcolor{my_c2}24.23 &
          \cellcolor{my_c2} \textbf{27.40} & \cellcolor{my_c2}\textbf{39.48}\\
    \bottomrule[1pt]
    \end{tabular}}
    \renewcommand{\arraystretch}{1} 
    \label{tab:result_llama2-7b}
\end{table*}

\textbf{Baselines.} To validate the effectiveness of \model~, we compared several structured pruning methods:
1) LLMPruner~\citep{ma2023llm}: Structural pruning via gradient-weight analysis to remove non-critical coupled layers; omits post-training for fair comparison but incurs pruning-policy overhead.
2) SliceGPT~\citep{ashkboos2024slicegpt}: PCA-based post-training sparsification reduces embedding dimensions by projecting hidden representations shallow-to-deep.
3) ShortGPT~\citep{men2024shortgpt}: Layer-pruning reveals redundancy in LLMs by removing redundant layers with minimal performance loss.
4) MHA-Drop~\citep{he2024matters}: Cosine-similarity-guided pruning of entire multi-head attention layers for inference acceleration.
5) FFN-Skip~\citep{jaiswal2024ffn}: Input-adaptive dynamic skipping of FFN layers during decoding for faster generation with negligible quality trade-offs. A detailed description of the baseline models can be found in Appendix \ref{appendix:benchmark}. \textit{Notably}, we diverge from previous works by evaluating all methods under identical memory budget, rather than a fixed pruning ratio. We posit that the pruning ratio is a misleading proxy for actual memory footprint, a claim substantiated by our empirical results which reveal a significant discrepancy. This gap arises primarily from the disproportionate memory overhead of runtime KV cache, which parameter counts alone fail to capture. By focusing on a fixed memory budget, our evaluation framework more accurately reflects the constraints of real-world deployment on resource-limited device, yielding more practical and reliable conclusions.

\subsection{Overall Performance}
\label{sec:results}

In this section, we evaluate LLama2-7B and LLama3-8B under 80\% and 60\% unified memory budgets, covering both parameters and KV cache. For clarity, 80\% memory budget corresponds to 80\% of the peak memory footprint of the original, unpruned model, formally expressed as $80\%*\text{max}(\text{param.}+\text{KV cache})$.  Sparsity is progressively increased for each method until the total memory overhead meets the target budget constraint. Detail pruning settings for all baselines are detailed in Table \ref{tab:remove_ratio} of Appendix~\ref{appendix:more_results}. Evaluated results with the same setting for Qwen1.5-7B and Qwen2.5-7B can be found in Table \ref{tab:appdenix_result} of Appendix~\ref{appendix:more_results}.

\noindent \textbf{Generation Ability.}
As shown in Table~\ref{tab:result_llama2-7b}, \model~ shows the smallest perplexity drift among all structured-pruning baselines. Specifically, at the 80\% budget perplexity rises on Llama2-7B by only  +6.33 on WikiText2 and  +21.47 on PTB, outperforming the next-best method by  22.95 and  146.93, respectively. This advantage persists more pronounced under the harsher 60\% cap, stemming from learning an architecturally-aware pruning policy targeting MHA when KV cache is the primary bottleneck and FFN blocks when parameter memory dominates. This asymmetric strategy adapts to the architectural nuances of different models. For instance, the FFN-heavy Llama3-8B (which uses GQA~\citep{ainslie2023gqa}) is highly sensitive to FFN removal, whereas the standard Llama2-7B is more degraded by pruning MHA. This learned selectivity allows \model~ to navigate architectural trade-offs, preserving crucial generative capabilities even under severe compression.

\noindent \textbf{Downstream Task Performance.}
We next evaluate zero-shot commonsense reasoning for the same memory budgets. As shown in Table~\ref{tab:result_llama2-7b}, \model~ again delivers the highest accuracy. Under the 80\% budget it preserves  ~86\% of dense accuracy and surpasses the leading baseline by  +8.2\% on Llama2-7B and  +11.5\% on Llama3-8B, with the largest gains on HellaSwag and ARC-e. Under an aggressive 60\% memory budget, while all methods degrade, \model~ proves uniquely resilient. It is the sole method to retain over  60\% of the dense model's performance, achieving scores of  43.0\% on Llama2-7B (0.6\% $\uparrow$ vs. runner-up) and  39.5\% on Llama3-8B (2.4\% $\uparrow$ vs. runner-up). These results indicate that \model~'s memory-aware, block-level pruning, which considers both parameter and KV cache memory constraints, provides superior performance retention compared to conventional approaches under severe memory limitations.

\begin{figure}[!t]
    \centering
    \includegraphics[width=0.75\linewidth]{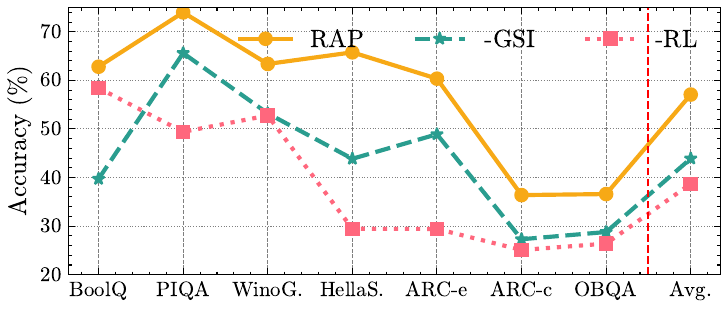}
    \includegraphics[width=0.75\linewidth]{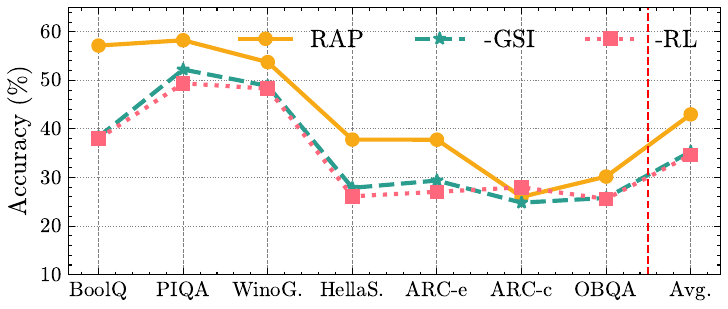}
    \caption{Effectiveness of GSI and RL Agent. Zero-shot performance of \model~$^{-\text{GSI}}$, \model~$^{-\text{RL}}$  versus \model~ evaluation on Llama2-7B under (a) 80\% and (b) 60\% memory budgets.}
    \label{fig:remove_GSI}
\end{figure}

\subsection{Ablation Study}

To explore the contribution of each component in \model~, we design two ablation variants: (1) \model~$^{-\text{GSI}}$, which disables the iterative, Greedy Sequential Importance scorer and instead applies standard static, one-shot perplexity scoring across all requests; and (2) \model~$^{-\text{RL}}$, which removes the RL agent and randomly drops blocks, where `$-$' means disable or remove proposed module.

\noindent \textbf{Effectiveness of GSI.}
To isolate the contribution of the Greedy Sequential Importance, we implement a static baseline that performs one-shot perplexity evaluation on all blocks initially, then removes the k blocks with the lowest importance to meet the memory budget, without iterative re-evaluation after each removal.
Figure \ref{fig:remove_GSI} and Table \ref{tab:remove_ppl} reveal that this shortcut severely erodes quality: perplexity on WikiText2 increases to 42.04 and average commonsense accuracy reduces by 13.17\%. 
The degradation arises from latent inter-block dependencies in transformer stacks. Conventional one-shot methods, which score each block independently within the context of the full model, produce misleadingly optimistic importance estimates. These estimates become invalid under multi-block pruning scenarios, as they fail to account for inter-block dependencies. GSI addresses this by iteratively pruning the least critical block and then recalibrating the importance of all remaining blocks within the new, contracted architecture. This sequential, state-aware evaluation yields more faithful importance scores, leading to superior performance in high compression regimes.

\begin{table}[t]
\centering
\caption{Ablation study on perplexity.}
\label{tab:remove_ppl}
\resizebox{0.75\linewidth}{!}{%
\begin{tabular}{c|lcc}
 \toprule
 \multirow{2}{*}{\textbf{Budget}} & \multirow{2}{*}{\textbf{Schemes}} & \multicolumn{2}{c}{\textbf{Perplexity} $\downarrow$} \\
  & & \textbf{WikiText2} & \textbf{PTB} \\
 \midrule
 \multirow{3}{*}{80\%} 
  & \textsc{Model}$^{-RL}$   & 313.51 & 535.75 \\
  & \textsc{Model}$^{-GSI}$  & 42.04  & 492.97 \\
  & \cellcolor{my_c2}\textsc{Model} & \cellcolor{my_c2}\textbf{11.80} & \cellcolor{my_c2}\textbf{46.56} \\
 \midrule
 \multirow{3}{*}{60\%} 
  & \textsc{Model}$^{-RL}$   & 7249.24 & 9059.14 \\
  & \textsc{Model}$^{-GSI}$  & 803.72  & 977.01 \\
  & \cellcolor{my_c2}\textsc{Model} & \cellcolor{my_c2}\textbf{74.78} & \cellcolor{my_c2}\textbf{592.65} \\
\bottomrule
\end{tabular}%
}
\end{table}

\paragraph{Effectiveness of RL Agent.}
We next ablate the RL agent that converts GSI scores into real-time actions by comparing \model~ with a naïve “Random-Drop” baseline that discards the same number of blocks but chooses them uniformly at random. As Figure \ref{fig:remove_GSI} and Table \ref{tab:remove_ppl} show, both variants satisfy the memory target, yet \model~ obviously exceeds the random baseline on generation ability ( -301.71 on WikiText2) and downstream performance ( +18.37\%). Crucially, \model~ accomplishes this online: its RL controller makes block-selection decisions conditioned on the current KV/parameter split, so the policy can tighten or relax MHA/FFN pruning as the request mix shifts. Random-Drop lacks such awareness; each inference call therefore risks violating latency or memory constraints on resource-constrained devices. In short, RL preserves GSI’s quality while adding workload-adaptive guarantees, making \model~ the more practical choice for on-device deployment.





\begin{figure}[!t]
    \centering
    \includegraphics[width=0.75\linewidth]{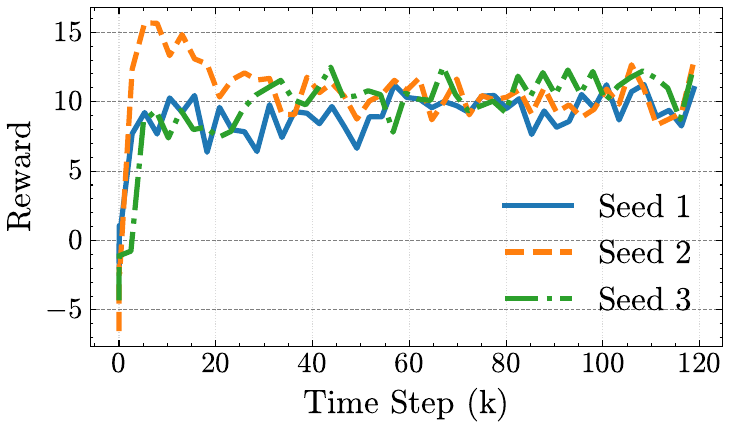}
    \caption{RL reward across different seeds.}
    \label{fig:reward}
\end{figure}

\subsection{Framework Analysis}
\textbf{Robustness of \model~.}

To verify that \model~’s learning dynamics are not brittle to initialization, we retrain the agent on Llama2-7B with three independent random seeds. Figure \ref{fig:reward} plots the seed-wise expected-reward curves. All trajectories increase smoothly and converge within a narrow band, showing that the agent consistently discovers high-quality pruning policies despite stochastic exploration. This stability stems from the architecture introduced above: (i) the Greedy Sequential Importance scorer supplies a well-shaped, low-variance reward signal, and (ii) the memory-aware action mask constrains the search space so early missteps cannot derail policy improvement. Collectively, these components make \model~’s reinforcement learning process both robust and generalizable across random seeds.

\begin{figure}[!t]
    \centering
    \includegraphics[width=0.8\linewidth]{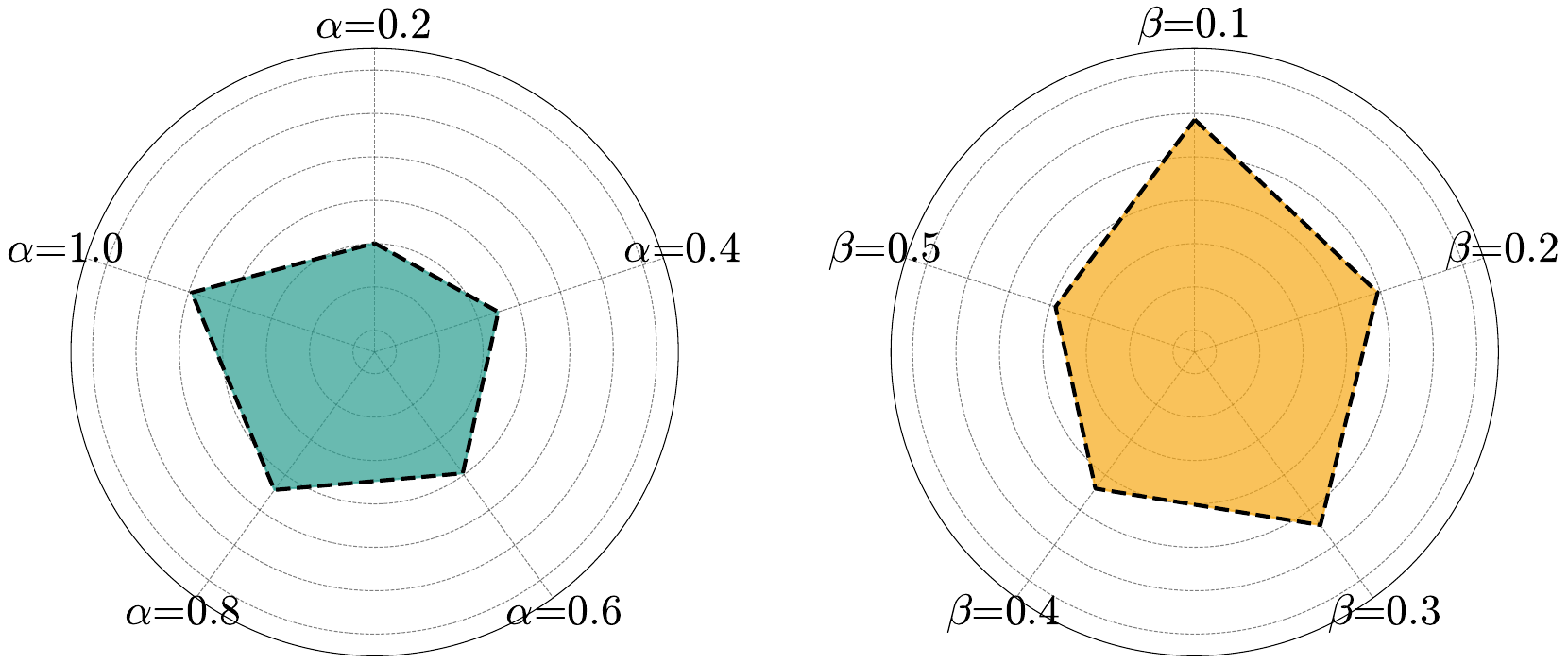}
    \caption{Penalty factors ($\alpha$ and $\beta$) sensitivity.}
    \label{fig:factors}
\end{figure}

\textbf{Impact of penalty factors \texorpdfstring{$\alpha$}{alpha} and \texorpdfstring{$\beta$}{beta}.}
\model~ reward function integrates task utility with two penalty terms, weighted by $\alpha$ and $\beta$, to discourage accuracy degradation (importance decay) and excessive memory usage, respectively. By sweeping $\alpha\in[0.2,1.0]$ and $\beta\in[0.1,0.5]$, users can tune the performance--efficiency trade-off to match deployment needs. As shown in Figure~\ref{fig:factors}, higher $\alpha$ values guide the policy to preserve critical blocks, while higher $\beta$ values encourage pruning memory-intensive ones. The optimal reward ridge emerges at large $\alpha$ and moderate $\beta$; we adopt $\alpha{=}1.0$ and $\beta{=}0.3$ in all experiments.

\textbf{Overhead Analysis.}
As shown in Figure \ref{fig:overhead} in Appendix~\ref{appendix:efficiency}. \model~'s RL controller adds negligible deployment overhead. While Llama2-7B has $\sim$6.7B parameters and requires 33GB memory for 2048-token inference at batch size 8, the controller has just 18K parameters over $3.7\times10^{5}\!\times$ reduction. Latency overhead is negligible: the unpruned model requires 52.73s for inference, whereas a policy step completes in 0.5s ( $<1\%$ overhead). Even including the one-time 302s offline policy training, the amortized cost is negligible. This efficiency stems from the controller’s compact two-layer MLP, which processes GSI scores and applies memory-aware masking to accelerate pruning. Furthermore, we analyze the computational cost of GSI and the end-to-end latency on edge devices. In practical deployment on resource-constrained edge devices, RAP achieves significantly lower latency and higher throughput compared to baselines (see Table~\ref{table:gsi_time} and Table~\ref{tab:latency&throughput} in Appendix~\ref{appendix:efficiency}).

\section{Conclusion}
\label{section6}
This paper addresses the deployment challenges of LLMs caused by their excessive computational and memory demands. While compression techniques have been proposed to mitigate these constraints, existing methods rely on static heuristics and fail to adapt to runtime memory fluctuations or heterogeneous KV cache requirements stemming from diverse user workloads. To overcome these limitations, we introduce \model~, an elastic pruning framework powered by RL that dynamically optimizes compression strategies in real-time based on system conditions. This work bridges the gap between static compression techniques and dynamic real-world deployment scenarios, offering a scalable solution for efficient LLM inference in heterogeneous environments. 
\section{Impact Statement}
\label{impact}

\paragraph{Positive Societal Impacts}
The proposed \model~ framework can substantially lower the computational and memory footprint of LLMs, thereby widening access for researchers, small enterprises, and under-resourced regions. By reducing both inference latency and energy consumption, \model~ facilitates the deployment of privacy-preserving on-device assistants and real-time multilingual tools in settings such as healthcare triage, disaster response, and education, where network connectivity or cloud resources are limited. Moreover, the fine-grained control over model capacity afforded by \model~ may accelerate scientific discovery: practitioners can iterate more rapidly on specialized models while retaining the interpretability of the original backbone, fostering reproducible and transparent AI research. Finally, by curbing redundant computation, the method contributes to a reduced carbon footprint for training and serving LLMs, aligning with global sustainability goals.

\paragraph{Negative Societal Impacts}
Conversely, the very efficiencies introduced by \model~ may inadvertently amplify risks associated with LLM proliferation. Lower hardware barriers could enable malicious actors to integrate compact yet potent language models into disinformation pipelines, automated phishing campaigns, or large-scale spam generation. The ability to tailor pruning policies to specific downstream tasks might also facilitate the creation of domain-targeted deepfakes or toxic content filters that evade existing moderation systems. In addition, the accelerated adoption of slimmed-down LLMs could exacerbate labor displacement in sectors reliant on routine text production or customer support. Finally, although \model~ decreases per-model energy usage, the ease of spawning numerous lightweight instances may lead to a rebound effect, offsetting aggregate environmental gains unless accompanied by conscientious governance and usage policies.


\bibliography{ref}

@inproceedings{frantar2023sparsegpt,
  title={Sparsegpt: Massive language models can be accurately pruned in one-shot},
  author={Frantar, Elias and Alistarh, Dan},
  booktitle={International Conference on Machine Learning},
  pages={10323--10337},
  year={2023},
  organization={PMLR}
}

@inproceedings{ma2023llmpruner,
  title={LLM-Pruner: On the Structural Pruning of Large Language Models},
  author={Xinyin Ma and Gongfan Fang and Xinchao Wang},
  booktitle={Advances in Neural Information Processing Systems},
  year={2023},
}

@misc{sun2024simpleeffectivepruningapproach,
      title={A Simple and Effective Pruning Approach for Large Language Models}, 
      author={Mingjie Sun and Zhuang Liu and Anna Bair and J. Zico Kolter},
      year={2024},
      eprint={2306.11695},
      archivePrefix={arXiv},
      primaryClass={cs.CL},
      url={https://arxiv.org/abs/2306.11695}, 
}

@inproceedings{shao2024one,
  title={One-shot sensitivity-aware mixed sparsity pruning for large language models},
  author={Shao, Hang and Liu, Bei and Qian, Yanmin},
  booktitle={ICASSP 2024-2024 IEEE International Conference on Acoustics, Speech and Signal Processing (ICASSP)},
  pages={11296--11300},
  year={2024},
  organization={IEEE}
}

@article{ma2023llm,
  title={Llm-pruner: On the structural pruning of large language models},
  author={Ma, Xinyin and Fang, Gongfan and Wang, Xinchao},
  journal={Advances in neural information processing systems},
  volume={36},
  pages={21702--21720},
  year={2023}
}

@article{touvron2023llama,
  title={Llama: Open and efficient foundation language models},
  author={Touvron, Hugo and Lavril, Thibaut and Izacard, Gautier and Martinet, Xavier and Lachaux, Marie-Anne and Lacroix, Timoth{\'e}e and Rozi{\`e}re, Baptiste and Goyal, Naman and Hambro, Eric and Azhar, Faisal and others},
  journal={arXiv preprint arXiv:2302.13971},
  year={2023}
}

@article{hu2021lora,
  title={Lora: Low-rank adaptation of large language models},
  author={Hu, Edward J and Shen, Yelong and Wallis, Phillip and Allen-Zhu, Zeyuan and Li, Yuanzhi and Wang, Shean and Wang, Lu and Chen, Weizhu},
  journal={arXiv preprint arXiv:2106.09685},
  year={2021}
}

@article{zhang2023loraprune,
  title={Loraprune: Pruning meets low-rank parameter-efficient fine-tuning},
  author={Zhang, Mingyang and Chen, Hao and Shen, Chunhua and Yang, Zhen and Ou, Linlin and Yu, Xinyi and Zhuang, Bohan},
  journal={arXiv preprint arXiv:2305.18403},
  year={2023}
}

@inproceedings{liu2023deja,
  title={Deja vu: Contextual sparsity for efficient llms at inference time},
  author={Liu, Zichang and Wang, Jue and Dao, Tri and Zhou, Tianyi and Yuan, Binhang and Song, Zhao and Shrivastava, Anshumali and Zhang, Ce and Tian, Yuandong and Re, Christopher and others},
  booktitle={International Conference on Machine Learning},
  pages={22137--22176},
  year={2023},
  organization={PMLR}
}

@article{zhong2024blockpruner,
  title={BlockPruner: Fine-grained Pruning for Large Language Models},
  author={Zhong, Longguang and Wan, Fanqi and Chen, Ruijun and Quan, Xiaojun and Li, Liangzhi},
  journal={arXiv preprint arXiv:2406.10594},
  year={2024}
}

@article{allenai:arc,
      author    = {Peter Clark  and Isaac Cowhey and Oren Etzioni and Tushar Khot and
                    Ashish Sabharwal and Carissa Schoenick and Oyvind Tafjord},
      title     = {Think you have Solved Question Answering? Try ARC, the AI2 Reasoning Challenge},
      journal   = {arXiv:1803.05457v1},
      year      = {2018},
}

@article{paszke2019pytorch,
  title={Pytorch: An imperative style, high-performance deep learning library},
  author={Paszke, Adam and Gross, Sam and Massa, Francisco and Lerer, Adam and Bradbury, James and Chanan, Gregory and Killeen, Trevor and Lin, Zeming and Gimelshein, Natalia and Antiga, Luca and others},
  journal={Advances in neural information processing systems},
  volume={32},
  year={2019}
}

@article{wolf2019huggingface,
  title={Huggingface's transformers: State-of-the-art natural language processing},
  author={Wolf, Thomas and Debut, Lysandre and Sanh, Victor and Chaumond, Julien and Delangue, Clement and Moi, Anthony and Cistac, Pierric and Rault, Tim and Louf, R{\'e}mi and Funtowicz, Morgan and others},
  journal={arXiv preprint arXiv:1910.03771},
  year={2019}
}

@article{touvron2023llama2,
  title={Llama 2: Open foundation and fine-tuned chat models},
  author={Touvron, Hugo and Martin, Louis and Stone, Kevin and Albert, Peter and Almahairi, Amjad and Babaei, Yasmine and Bashlykov, Nikolay and Batra, Soumya and Bhargava, Prajjwal and Bhosale, Shruti and others},
  journal={arXiv preprint arXiv:2307.09288},
  url={https://arxiv.org/abs/2307.09288},
  year={2023}
}

@misc{eval-harness,
  author       = {Gao, Leo and Tow, Jonathan and Abbasi, Baber and Biderman, Stella and Black, Sid and DiPofi, Anthony and Foster, Charles and Golding, Laurence and Hsu, Jeffrey and Le Noac'h, Alain and Li, Haonan and McDonell, Kyle and Muennighoff, Niklas and Ociepa, Chris and Phang, Jason and Reynolds, Laria and Schoelkopf, Hailey and Skowron, Aviya and Sutawika, Lintang and Tang, Eric and Thite, Anish and Wang, Ben and Wang, Kevin and Zou, Andy},
  title        = {A framework for few-shot language model evaluation},
  month        = 12,
  year         = 2023,
  publisher    = {Zenodo},
  version      = {v0.4.0},
  doi          = {10.5281/zenodo.10256836},
  url          = {https://zenodo.org/records/10256836}
}

@article{brown2020language,
  title={Language models are few-shot learners},
  author={Brown, Tom and Mann, Benjamin and Ryder, Nick and Subbiah, Melanie and Kaplan, Jared D and Dhariwal, Prafulla and Neelakantan, Arvind and Shyam, Pranav and Sastry, Girish and Askell, Amanda and others},
  journal={Advances in neural information processing systems},
  volume={33},
  pages={1877--1901},
  year={2020}
}

@article{achiam2023gpt,
  title={Gpt-4 technical report},
  author={Achiam, Josh and Adler, Steven and Agarwal, Sandhini and Ahmad, Lama and Akkaya, Ilge and Aleman, Florencia Leoni and Almeida, Diogo and Altenschmidt, Janko and Altman, Sam and Anadkat, Shyamal and others},
  journal={arXiv preprint arXiv:2303.08774},
  year={2023}
}

@misc{microsoft_bing_new_features,
  author ={microsoft},
  title = {{Your Everyday AI Companion | Microsoft Bing}},
  howpublished = {\url{https://www.bing.com/new}},
  organization = {Microsoft}
}

@misc{github_copilot,
 author = {github},
  title = {{GitHub Copilot: Your AI pair programmer}},
  howpublished = {\url{https://github.com/features/copilot}},
  organization = {GitHub}
}

@article{fedus2022review,
  title={A review of sparse expert models in deep learning},
  author={Fedus, William and Dean, Jeff and Zoph, Barret},
  journal={arXiv preprint arXiv:2209.01667},
  year={2022}
}

@article{patterson2021carbon,
  title={Carbon emissions and large neural network training},
  author={Patterson, David and Gonzalez, Joseph and Le, Quoc and Liang, Chen and Munguia, Lluis-Miquel and Rothchild, Daniel and So, David and Texier, Maud and Dean, Jeff},
  journal={arXiv preprint arXiv:2104.10350},
  year={2021}
}

@article{llama_2,
  title={Llama 2: Open foundation and fine-tuned chat models},
  author={Touvron, Hugo and Martin, Louis and Stone, Kevin and Albert, Peter and Almahairi, Amjad and Babaei, Yasmine and Bashlykov, Nikolay and Batra, Soumya and Bhargava, Prajjwal and Bhosale, Shruti and others},
  journal={arXiv preprint arXiv:2307.09288},
  year={2023}
}

@article{PaLM,
  author       = {Aakanksha Chowdhery and
                  Sharan Narang and
                  Jacob Devlin},
  title        = {PaLM: Scaling Language Modeling with Pathways},
  journal      = {J. Mach. Learn. Res.},
  volume       = {24},
  pages        = {240:1--240:113},
  year         = {2023},
  url          = {http://jmlr.org/papers/v24/22-1144.html},
  bibsource    = {dblp computer science bibliography, https://dblp.org}
}

@article{gemma,
  title={Gemma: Open models based on gemini research and technology},
  author={Team, Gemma and Mesnard, Thomas and Hardin, Cassidy and Dadashi, Robert and Bhupatiraju, Surya and Pathak, Shreya and Sifre, Laurent and Rivi{\`e}re, Morgane and Kale, Mihir Sanjay and Love, Juliette and others},
  journal={arXiv preprint arXiv:2403.08295},
  year={2024}
}

@article{yuan2023mobile,
  title={Mobile Foundation Model as Firmware},
  author={Yuan, Jinliang and Yang, Chen and Cai, Dongqi and Wang, Shihe and Yuan, Xin and Zhang, Zeling and Li, Xiang and Zhang, Dingge and Mei, Hanzi and Jia, Xianqing and others},
  journal={arXiv preprint arXiv:2308.14363},
  year={2023}
}

@article{lin2022device,
  title={On-device training under 256kb memory},
  author={Lin, Ji and Zhu, Ligeng and Chen, Wei-Ming and Wang, Wei-Chen and Gan, Chuang and Han, Song},
  journal={Advances in Neural Information Processing Systems},
  volume={35},
  pages={22941--22954},
  year={2022}
}

@misc{alpaca,
  author = {Rohan Taori and Ishaan Gulrajani and Tianyi Zhang and Yann Dubois and Xuechen Li and Carlos Guestrin and Percy Liang and Tatsunori B. Hashimoto },
  title = {Stanford Alpaca: An Instruction-following LLaMA model},
  year = {2023},
  publisher = {GitHub},
  journal = {GitHub repository},
  howpublished = {\url{https://github.com/tatsu-lab/stanford_alpaca}},
}

@article{dubey2024llama,
  title={The llama 3 herd of models},
  author={Dubey, Abhimanyu and Jauhri, Abhinav and Pandey, Abhinav and Kadian, Abhishek and Al-Dahle, Ahmad and Letman, Aiesha and Mathur, Akhil and Schelten, Alan and Yang, Amy and Fan, Angela and others},
  journal={arXiv preprint arXiv:2407.21783},
  year={2024}
}

@article{yang2024qwen2,
  title={Qwen2 technical report},
  author={Yang, An and Yang, Baosong and Hui, Binyuan and Zheng, Bo and Yu, Bowen and Zhou, Chang and Li, Chengpeng and Li, Chengyuan and Liu, Dayiheng and Huang, Fei and others},
  journal={arXiv preprint arXiv:2407.10671},
  year={2024}
}

@article{ainslie2023gqa,
  title={Gqa: Training generalized multi-query transformer models from multi-head checkpoints},
  author={Ainslie, Joshua and Lee-Thorp, James and de Jong, Michiel and Zemlyanskiy, Yury and Lebr{\'o}n, Federico and Sanghai, Sumit},
  journal={arXiv preprint arXiv:2305.13245},
  year={2023}
}

@inproceedings{Bisk2020_piqa,
  author = {Yonatan Bisk and Rowan Zellers and Ronan Le Bras and Jianfeng Gao and Yejin Choi},
  title = {PIQA: Reasoning about Physical Commonsense in Natural Language},
  booktitle = {Thirty-Fourth AAAI Conference on Artificial Intelligence},
  year = {2020},
}

@inproceedings{zellers2019hellaswag,
    title={HellaSwag: Can a Machine Really Finish Your Sentence?},
    author={Zellers, Rowan and Holtzman, Ari and Bisk, Yonatan and Farhadi, Ali and Choi, Yejin},
    booktitle ={Proceedings of the 57th Annual Meeting of the Association for Computational Linguistics},
    year={2019}
}

@misc{sakaguchi2019winograndeadversarialwinogradschema,
      title={WinoGrande: An Adversarial Winograd Schema Challenge at Scale}, 
      author={Keisuke Sakaguchi and Ronan Le Bras and Chandra Bhagavatula and Yejin Choi},
      year={2019},
      eprint={1907.10641},
      archivePrefix={arXiv},
      primaryClass={cs.CL},
      url={https://arxiv.org/abs/1907.10641}, 
}

@inproceedings{OpenBookQA2018,
 title={Can a Suit of Armor Conduct Electricity? A New Dataset for Open Book Question Answering},
 author={Todor Mihaylov and Peter Clark and Tushar Khot and Ashish Sabharwal},
 booktitle={EMNLP},
 year={2018}
}

@misc{clark2019boolqexploringsurprisingdifficulty,
      title={BoolQ: Exploring the Surprising Difficulty of Natural Yes/No Questions}, 
      author={Christopher Clark and Kenton Lee and Ming-Wei Chang and Tom Kwiatkowski and Michael Collins and Kristina Toutanova},
      year={2019},
      eprint={1905.10044},
      archivePrefix={arXiv},
      primaryClass={cs.CL},
      url={https://arxiv.org/abs/1905.10044}, 
}

@article{ashkboos2024slicegpt,
  title={Slicegpt: Compress large language models by deleting rows and columns},
  author={Ashkboos, Saleh and Croci, Maximilian L and Nascimento, Marcelo Gennari do and Hoefler, Torsten and Hensman, James},
  journal={arXiv preprint arXiv:2401.15024},
  year={2024}
}

@article{ling2024slimgpt,
  title={SlimGPT: Layer-wise Structured Pruning for Large Language Models},
  author={Ling, Gui and Wang, Ziyang and Yan, Yuliang and Liu, Qingwen},
  journal={arXiv preprint arXiv:2412.18110v1},
  year={2024},
  url={https://arxiv.org/abs/2412.18110v1}
}

@inproceedings{meng2024osscar,
  title={OSSCAR: One-Shot Structured Pruning in Vision and Language Models with Combinatorial Optimization},
  author={Meng, Xiang and Ibrahim, Shibal and Behdin, Kayhan and Hazimeh, Hussein and Ponomareva, Natalia and Mazumder, Rahul},
  booktitle={Proceedings of the 41st International Conference on Machine Learning},
  pages={325--335},
  year={2024},
  organization={PMLR},
  url={https://openreview.net/pdf?id=ZctlF8RlV4}
}

@inproceedings{trace,
  author       = {Jovan Stojkovic and
                  Chaojie Zhang and
                  {\'{I}}{\~{n}}igo Goiri and
                  Josep Torrellas and
                  Esha Choukse},
  title        = {DynamoLLM: Designing {LLM} Inference Clusters for Performance and
                  Energy Efficiency},
  booktitle    = {{IEEE} International Symposium on High Performance Computer Architecture,
                  {HPCA} 2025, Las Vegas, NV, USA, March 1-5, 2025},
  pages        = {1348--1362},
  publisher    = {{IEEE}},
  year         = {2025},
  url          = {https://doi.org/10.1109/HPCA61900.2025.00102},
  doi          = {10.1109/HPCA61900.2025.00102},
  bibsource    = {dblp computer science bibliography, https://dblp.org}
}

@inproceedings{patel2024splitwise,
  title={Splitwise: Efficient generative llm inference using phase splitting},
  author={Patel, Pratyush and Choukse, Esha and Zhang, Chaojie and Shah, Aashaka and Goiri, {\'I}{\~n}igo and Maleki, Saeed and Bianchini, Ricardo},
  booktitle={2024 ACM/IEEE 51st Annual International Symposium on Computer Architecture (ISCA)},
  pages={118--132},
  year={2024},
  organization={IEEE}
}

@article{li2024llm,
  title={Llm inference serving: Survey of recent advances and opportunities},
  author={Li, Baolin and Jiang, Yankai and Gadepally, Vijay and Tiwari, Devesh},
  journal={arXiv preprint arXiv:2407.12391},
  year={2024}
}

@article{jaiswal2025serving,
  title={Serving models, fast and slow: optimizing heterogeneous LLM inferencing workloads at scale},
  author={Jaiswal, Shashwat and Jain, Kunal and Simmhan, Yogesh and Parayil, Anjaly and Mallick, Ankur and Wang, Rujia and Amant, Renee St and Bansal, Chetan and R{\"u}hle, Victor and Kulkarni, Anoop and others},
  journal={arXiv preprint arXiv:2502.14617},
  year={2025}
}

@article{zhang2024investigating,
  title={Investigating layer importance in large language models},
  author={Zhang, Yang and Dong, Yanfei and Kawaguchi, Kenji},
  journal={arXiv preprint arXiv:2409.14381},
  year={2024}
}

@article{he2024matters,
  title={What matters in transformers? not all attention is needed},
  author={He, Shwai and Sun, Guoheng and Shen, Zheyu and Li, Ang},
  journal={arXiv preprint arXiv:2406.15786},
  year={2024}
}

@article{yao2024layer,
  title={Layer-wise Importance Matters: Less Memory for Better Performance in Parameter-efficient Fine-tuning of Large Language Models},
  author={Yao, Kai and Gao, Penglei and Li, Lichun and Zhao, Yuan and Wang, Xiaofeng and Wang, Wei and Zhu, Jianke},
  journal={arXiv preprint arXiv:2410.11772},
  year={2024}
}

@article{pan2024lisa,
  title={LISA: layerwise importance sampling for memory-efficient large language model fine-tuning},
  author={Pan, Rui and Liu, Xiang and Diao, Shizhe and Pi, Renjie and Zhang, Jipeng and Han, Chi and Zhang, Tong},
  journal={Advances in Neural Information Processing Systems},
  volume={37},
  pages={57018--57049},
  year={2024}
}

@article{men2024shortgpt,
  title={Shortgpt: Layers in large language models are more redundant than you expect},
  author={Men, Xin and Xu, Mingyu and Zhang, Qingyu and Wang, Bingning and Lin, Hongyu and Lu, Yaojie and Han, Xianpei and Chen, Weipeng},
  journal={arXiv preprint arXiv:2403.03853},
  year={2024}
}

@article{jaiswal2024ffn,
  title={Ffn-skipllm: A hidden gem for autoregressive decoding with adaptive feed forward skipping},
  author={Jaiswal, Ajay and Hu, Bodun and Yin, Lu and Ro, Yeonju and Liu, Shiwei and Chen, Tianlong and Akella, Aditya},
  journal={arXiv preprint arXiv:2404.03865},
  year={2024}
}

@article{gao2024disp,
  title={Disp-llm: Dimension-independent structural pruning for large language models},
  author={Gao, Shangqian and Lin, Chi-Heng and Hua, Ting and Tang, Zheng and Shen, Yilin and Jin, Hongxia and Hsu, Yen-Chang},
  journal={Advances in Neural Information Processing Systems},
  volume={37},
  pages={72219--72244},
  year={2024}
}

@article{wang2024burstgpt,
  title={BurstGPT: A Real-world Workload Dataset to Optimize LLM Serving Systems},
  author={Wang, Yuxin and Chen, Yuhan and Li, Zeyu and Kang, Xueze and Tang, Zhenheng and He, Xin and Guo, Rui and Wang, Xin and Wang, Qiang and Zhou, Amelie Chi and others},
  journal={arXiv preprint arXiv:2401.17644},
  year={2024}
}

@article{yu2023faaswap,
  title={Faaswap: slo-aware, gpu-efficient serverless inference via model swapping},
  author={Yu, Minchen and Wang, Ao and Chen, Dong and Yu, Haoxuan and Luo, Xiaonan and Li, Zhuohao and Wang, Wei and Chen, Ruichuan and Nie, Dapeng and Yang, Haoran},
  journal={arXiv preprint arXiv:2306.03622},
  year={2023}
}

@article{xu2022igniter,
  title={igniter: Interference-aware gpu resource provisioning for predictable dnn inference in the cloud},
  author={Xu, Fei and Xu, Jianian and Chen, Jiabin and Chen, Li and Shang, Ruitao and Zhou, Zhi and Liu, Fangming},
  journal={IEEE Transactions on Parallel and Distributed Systems},
  volume={34},
  number={3},
  pages={812--827},
  year={2022},
  publisher={IEEE}
}

@misc{nvidia2020a40,
  author       = {{NVIDIA Corporation}},
  title        = {{NVIDIA A40 GPU}},
  howpublished = {\url{https://www.nvidia.com/en-us/data-center/a40/}},
  year         = {2020}
}

@misc{merity2016pointer,
      title={Pointer Sentinel Mixture Models},
      author={Stephen Merity and Caiming Xiong and James Bradbury and Richard Socher},
      year={2016},
      eprint={1609.07843},
      archivePrefix={arXiv},
      primaryClass={cs.CL}
}

@article{marcus1993building,
  title={Building a large annotated corpus of English: The Penn Treebank},
  author={Marcus, Mitch and Santorini, Beatrice and Marcinkiewicz, Mary Ann},
  journal={Computational linguistics},
  volume={19},
  number={2},
  pages={313--330},
  year={1993}
}

@article{sun2019patient,
  title={Patient knowledge distillation for bert model compression},
  author={Sun, Siqi and Cheng, Yu and Gan, Zhe and Liu, Jingjing},
  journal={arXiv preprint arXiv:1908.09355},
  year={2019}
}

@article{xu2024survey,
  title={A survey on knowledge distillation of large language models},
  author={Xu, Xiaohan and Li, Ming and Tao, Chongyang and Shen, Tao and Cheng, Reynold and Li, Jinyang and Xu, Can and Tao, Dacheng and Zhou, Tianyi},
  journal={arXiv preprint arXiv:2402.13116},
  year={2024}
}

@article{chen2024bge,
  title={Bge m3-embedding: Multi-lingual, multi-functionality, multi-granularity text embeddings through self-knowledge distillation},
  author={Chen, Jianlv and Xiao, Shitao and Zhang, Peitian and Luo, Kun and Lian, Defu and Liu, Zheng},
  journal={arXiv preprint arXiv:2402.03216},
  year={2024}
}

@article{liu2024spinquant,
  title={Spinquant: Llm quantization with learned rotations},
  author={Liu, Zechun and Zhao, Changsheng and Fedorov, Igor and Soran, Bilge and Choudhary, Dhruv and Krishnamoorthi, Raghuraman and Chandra, Vikas and Tian, Yuandong and Blankevoort, Tijmen},
  journal={arXiv preprint arXiv:2405.16406},
  year={2024}
}

@article{lin2024awq,
  title={Awq: Activation-aware weight quantization for on-device llm compression and acceleration},
  author={Lin, Ji and Tang, Jiaming and Tang, Haotian and Yang, Shang and Chen, Wei-Ming and Wang, Wei-Chen and Xiao, Guangxuan and Dang, Xingyu and Gan, Chuang and Han, Song},
  journal={Proceedings of Machine Learning and Systems},
  volume={6},
  pages={87--100},
  year={2024}
}

@inproceedings{an2024fluctuation,
  title={Fluctuation-based adaptive structured pruning for large language models},
  author={An, Yongqi and Zhao, Xu and Yu, Tao and Tang, Ming and Wang, Jinqiao},
  booktitle={Proceedings of the AAAI Conference on Artificial Intelligence},
  volume={38},
  pages={10865--10873},
  year={2024}
}

@article{federici2024efficient,
  title={Efficient LLM Inference using Dynamic Input Pruning and Cache-Aware Masking},
  author={Federici, Marco and Belli, Davide and van Baalen, Mart and Jalalirad, Amir and Skliar, Andrii and Major, Bence and Nagel, Markus and Whatmough, Paul},
  journal={arXiv preprint arXiv:2412.01380},
  year={2024}
}

@article{le2025probe,
  title={Probe pruning: Accelerating llms through dynamic pruning via model-probing},
  author={Le, Qi and Diao, Enmao and Wang, Ziyan and Wang, Xinran and Ding, Jie and Yang, Li and Anwar, Ali},
  journal={arXiv preprint arXiv:2502.15618},
  year={2025}
}

@article{li2023sparse,
  title={E-sparse: Boosting the large language model inference through entropy-based n: M sparsity},
  author={Li, Yun and Niu, Lin and Zhang, Xipeng and Liu, Kai and Zhu, Jianchen and Kang, Zhanhui},
  journal={arXiv preprint arXiv:2310.15929},
  year={2023}
}

@article{das2023beyond,
  title={Beyond size: How gradients shape pruning decisions in large language models},
  author={Das, Rocktim Jyoti and Sun, Mingjie and Ma, Liqun and Shen, Zhiqiang},
  journal={arXiv preprint arXiv:2311.04902},
  year={2023}
}

@article{zhao2024apt,
  title={Apt: Adaptive pruning and tuning pretrained language models for efficient training and inference},
  author={Zhao, Bowen and Hajishirzi, Hannaneh and Cao, Qingqing},
  journal={arXiv preprint arXiv:2401.12200},
  year={2024}
}

@article{chen2023lorashear,
  title={Lorashear: Efficient large language model structured pruning and knowledge recovery},
  author={Chen, Tianyi and Ding, Tianyu and Yadav, Badal and Zharkov, Ilya and Liang, Luming},
  journal={arXiv preprint arXiv:2310.18356},
  year={2023}
}

@article{yin2023outlier,
  title={Outlier weighed layerwise sparsity (owl): A missing secret sauce for pruning llms to high sparsity},
  author={Yin, Lu and Wu, You and Zhang, Zhenyu and Hsieh, Cheng-Yu and Wang, Yaqing and Jia, Yiling and Li, Gen and Jaiswal, Ajay and Pechenizkiy, Mykola and Liang, Yi and others},
  journal={arXiv preprint arXiv:2310.05175},
  year={2023}
}

@article{xia2023flash,
  title={Flash-llm: Enabling cost-effective and highly-efficient large generative model inference with unstructured sparsity},
  author={Xia, Haojun and Zheng, Zhen and Li, Yuchao and Zhuang, Donglin and Zhou, Zhongzhu and Qiu, Xiafei and Li, Yong and Lin, Wei and Song, Shuaiwen Leon},
  journal={arXiv preprint arXiv:2309.10285},
  year={2023}
}

@article{zhang2024finercut,
  title={Finercut: Finer-grained interpretable layer pruning for large language models},
  author={Zhang, Yang and Li, Yawei and Wang, Xinpeng and Shen, Qianli and Plank, Barbara and Bischl, Bernd and Rezaei, Mina and Kawaguchi, Kenji},
  journal={arXiv preprint arXiv:2405.18218},
  year={2024}
}

@article{mnih2015human,
  title={Human-level control through deep reinforcement learning},
  author={Mnih, Volodymyr and Kavukcuoglu, Koray and Silver, David and Rusu, Andrei A and Veness, Joel and Bellemare, Marc G and Graves, Alex and Riedmiller, Martin and Fidjeland, Andreas K and Ostrovski, Georg and others},
  journal={nature},
  volume={518},
  number={7540},
  pages={529--533},
  year={2015},
  publisher={Nature Publishing Group}
}

@article{zhang2017deep,
  title={Deep reinforcement learning for visual object tracking in videos},
  author={Zhang, Da and Maei, Hamid and Wang, Xin and Wang, Yuan-Fang},
  journal={arXiv preprint arXiv:1701.08936},
  year={2017}
}

@article{andrychowicz2020learning,
  title={Learning dexterous in-hand manipulation},
  author={Andrychowicz, OpenAI: Marcin and Baker, Bowen and Chociej, Maciek and Jozefowicz, Rafal and McGrew, Bob and Pachocki, Jakub and Petron, Arthur and Plappert, Matthias and Powell, Glenn and Ray, Alex and others},
  journal={The International Journal of Robotics Research},
  volume={39},
  number={1},
  pages={3--20},
  year={2020},
  publisher={SAGE Publications Sage UK: London, England}
}

@article{bai2023qwen,
  title={Qwen technical report},
  author={Bai, Jinze and Bai, Shuai and Chu, Yunfei and Cui, Zeyu and Dang, Kai and Deng, Xiaodong and Fan, Yang and Ge, Wenbin and Han, Yu and Huang, Fei and others},
  journal={arXiv preprint arXiv:2309.16609},
  year={2023}
}
\bibliographystyle{icml2026}

\newpage
\appendix
\onecolumn
\section{Detail of RL-Agent Algorithm}
\label{appendix:Algorithm}

\subsection{Problem Formulation}

We cast RAP as a finite-horizon MDP $\mathcal{M}=(\mathcal S,\mathcal A,\mathcal P,\mathcal R,\gamma)$ with horizon $H\!\le\!2N$, where $N$ is the number of transformer layers and each layer contributes one MHA block and one FFN block (thus $2N$ removable blocks).

\paragraph{State.}
At decision step $t$, the state $s_t\in\mathcal S$ concatenates request-, model-, and system-level features:
\[
s_t \;=\; \big(s^{\text{Req}}_t,\, s^{\text{Model}}_t,\, s^{\text{Sys}}_t\big),
\]
with
\[
s^{\text{Req}}_t=(R_{\mathrm{bs}},\,R_{\mathrm{sql}}),\qquad
s^{\text{Model}}_t=\big(\{\mathrm{MHA}_{\mathrm{imp},i}^{(t)}\}_{i=1}^N,\{\mathrm{FFN}_{\mathrm{imp},i}^{(t)}\}_{i=1}^N\big),
\]
\[
s^{\text{Sys}}_t=(\mathrm{Sys}_{\mathrm{avail}}^{(t)},\,\widehat{\mathrm{Sys}}_{\mathrm{req}}^{(t)}).
\]
Here $\mathrm{MHA}_{\mathrm{imp},i}^{(t)}$ and $\mathrm{FFN}_{\mathrm{imp},i}^{(t)}$ are the \emph{current} Greedy Sequential Importance (GSI) scores recomputed after each removal (see Alg.~\ref{alg:GSI}); $\mathrm{Sys}_{\mathrm{avail}}^{(t)}$ is the available GPU memory observed at time $t$; and $\widehat{\mathrm{Sys}}_{\mathrm{req}}^{(t)}$ is the agent's estimate of the peak memory after applying the candidate action.

\paragraph{Action.}
We adopt \emph{sequential single-block} decisions compatible with DQN:
\[
\mathcal A=\{0,1,2,\dots,2N\}.
\]
Action $a_t=0$ denotes \textsc{Stop}; $a_t\in\{1,\dots,2N\}$ removes the corresponding block (one of the $N$ MHA or $N$ FFN blocks). An \emph{action mask} invalidates pruned blocks and can optionally disable actions predicted to break correctness constraints. The episode terminates when either: (i) \textsc{Stop} is taken, or (ii) the peak memory fits the budget.

\paragraph{Transition.}
Given $(s_t,a_t)$, the environment deterministically updates the pruned architecture $\mathcal M_t\!\mapsto\!\mathcal M_{t+1}$ by excising the selected block if $a_t\!>\!0$, then re-evaluates the GSI scores on the contracted model to produce $s_{t+1}$. Runtime memory availability $\mathrm{Sys}_{\mathrm{avail}}^{(t+1)}$ can be treated as exogenous.

\paragraph{Discount.}
We set $\gamma=0.99$.

\subsection{Memory Model (Peak GPU Footprint)}
Consistent with the main text, the peak inference memory comprises \emph{static parameters} and \emph{dynamic KV cache}.
Let $b_{\text{prec}}$ be bytes per scalar (e.g., 2 for bfloat16). For a model state $\mathcal M$ (after some blocks are removed) and a request tuple $(R_{\mathrm{bs}},R_{\mathrm{sql}})$, we estimate
\begin{align}
\mathrm{Mem}_{\text{param}}(\mathcal M) &= b_{\text{prec}} \sum_{B\in\mathcal B(\mathcal M)} \#\text{params}(B), \\
\mathrm{Mem}_{\text{KV}}(\mathcal M,R_{\mathrm{bs}},R_{\mathrm{sql}}) &= b_{\text{prec}} \cdot 2 \sum_{\ell\in\mathcal L(\mathcal M)} n_{\mathrm{heads},\ell} \, d_{\mathrm{head},\ell}\, R_{\mathrm{bs}}\, R_{\mathrm{sql}} ,
\end{align}
where $\mathcal B(\mathcal M)$ and $\mathcal L(\mathcal M)$ denote remaining blocks and layers, respectively; the factor $2$ stores keys and values. The peak is
\[
\mathrm{Mem}_{\text{peak}}(\mathcal M,R_{\mathrm{bs}},R_{\mathrm{sql}})=
\mathrm{Mem}_{\text{param}}(\mathcal M)+\mathrm{Mem}_{\text{KV}}(\mathcal M,R_{\mathrm{bs}},R_{\mathrm{sql}}).
\]
This matches the linear KV-cache scaling with batch and sequence length emphasized in the main paper.



\subsection{DQN-based Policy Learning with Action Masking}
Let $Q_\theta(s,a)$ be the action-value function and $Q_{\bar\theta}$ its target copy. We adopt masked $\varepsilon$-greedy:
\[
\pi(a|s) = 
\begin{cases}
\text{uniform over valid actions} & \text{with prob. } \varepsilon,\\
\arg\max_{a\in\mathcal A_{\text{valid}}(s)} Q_\theta(s,a) & \text{with prob. } 1-\varepsilon,
\end{cases}
\]
where $\mathcal A_{\text{valid}}(s)$ removes already-pruned blocks and can optionally include feasibility heuristics. With transitions $(s_t,a_t,r_t,s_{t+1},\text{done})$ stored in replay buffer $\mathcal D$, we minimize
\[
\mathcal L(\theta)=\mathbb E_{(s,a,r,s',d)\sim\mathcal D}\Big[\big(Q_\theta(s,a)-y\big)^2\Big],\quad
y=r+\gamma(1-d)\max_{a'\in\mathcal A_{\text{valid}}(s')} Q_{\bar\theta}(s',a').
\]
We soft-update the target network periodically: $\bar\theta\leftarrow\tau\theta+(1-\tau)\bar\theta$.

\section{Datasets and Baselines}

\subsection{Commonsene Reasoning}
\label{appendix:dataset}
The details of the benchmarks are as follows:

\begin{itemize}
    \item BoolQ~\citep{clark2019boolqexploringsurprisingdifficulty}: yes/no questions which are naturally occurring and generated in unprompted and unconstrained settings. There are 3270 questions in the test set.
    \item PIQA~\citep{Bisk2020_piqa}: questions with two solutions requiring physical commonsense. There are 1830 questions in the test set.
    \item HellaSwag~\citep{zellers2019hellaswag}: commonsense NLI questions including a context and several endings which complete the context. There are 10042 questions in the test set.
    \item WinoGrande~\citep{sakaguchi2019winograndeadversarialwinogradschema}: fill-in-a-blank task with binary options to choose the right option for a given sentence which requires commonsense reasoning. There are 1267 questions in the test set.
    \item ARC-easy~\citep{allenai:arc} \& ARC-challenge~\citep{allenai:arc}: the Challenge Set and Easy Set of ARC dataset of genuine grade-school level, containing 2376/1172 multiple-choice science questions in the test set, respectively.
    \item OpenbookQA~\citep{OpenBookQA2018}: uestions requiring multi-step reasoning, use of additional commonsense knowledge, and rich text comprehension. There are 500 questions in the test set.
\end{itemize}

\subsection{Baselines}
\label{appendix:benchmark}
\begin{itemize}
    \item \textit{LLMPruner}~\citep{ma2023llm}, which  adopts structural pruning that selectively removes non-critical coupled structures based on weights and gradient information, maximally preserving the majority of the LLM’s functionality. LLMPruner applies post training to the pruned model, but for fair comparison, we do not apply post training to it. However, LLMPuner requires extra overhead for pruning its gradient-base pruning policy.
    \item \textit{SliceGPT}~\citep{ashkboos2024slicegpt}, which is a post-training sparsification scheme that replaces each weight matrix with a smaller matrix, reducing the embedding dimension of the network. Specifically, they applied PCA to the hidden representation from shallow to deep layers, and incorporated the dimension reduction matrix into existing network parameters.
    \item \textit{DISP-LLM}~\citep{gao2024disp}, which introduces a dimension-independent structural pruning scheme that breaks inter-layer width coupling. This post-training method uses gradient-based optimization via a learned hyper-network to determine which neurons to remove in each layer, enabling flexible layer-specific width reduction without additional fine-tuning.
    \item \textit{ShortGPT}~\citep{men2024shortgpt}  reveals significant redundancy among LLMs by proposing a layer-pruning method that removes redundant layers with minimal performance degradation
    \item \textit{MHA-Drop}~\citep{he2024matters}, which prunes entire multi-head self-attention layers of Transformer blocks to accelerate inference. By removing a fraction of the attention layers based on cosine similarity-based importance, this approach achieves notable speedups with minor impact on the model performance.
    \item \textit{FFN-Skip}~\citep{jaiswal2024ffn}, which applys inference-time skipping strategy that omits selected feed-forward network layers to reduce computation. It leverages an input-adaptive criterion to dynamically skip FFN blocks during decoding, yielding faster generation with negligible degradation in output quality.
\end{itemize}

\section{More Results}
\label{appendix:more_results}
Table~\ref{tab:appdenix_result} shows additional results on Qwen-1.5-7B and Qwen-2.5-7B, which confirms the proposed \model~ is architecture-agnostic: it preserves competitive perplexity and downstream accuracy across two distinct generations of the Qwen series, implying that the same pruning strategy can be ported to other modern transformer backbones with minimal modification.

\begin{table*}[!t]
    \centering
    \caption{Zero‑shot performance of pruned versus dense model under different memory budgets. $^{1}$ 100\% memory budget indicates exceeding peak inference memory usage (parameters + KV cache).}
    \resizebox{\linewidth}{!}{\begin{tabular}{cc|cc|ccccccc|c}
    \toprule [1pt]
     \multirow{2}{*}{\textbf{
     Budget}} & \multirow{2}{*}{\textbf{Schemes}}   & \multicolumn{2}{c|}{\textbf{Perplexity} $\downarrow$} & \multicolumn{8}{c}{\textbf{Commonsense Task (\%)} $\uparrow$} \\ 
         & & \textbf{WikiText2} & \textbf{PTB}& \textbf{BoolQ} &\textbf{PIQA} & \textbf{WinoG.} & \textbf{HellaS.} & \textbf{ARC-e} & \textbf{ARC-c} & \textbf{OBQA} & \textbf{Avg.} \\ \midrule [1pt]
      \rowcolor{gray!10}   \multicolumn{12}{c}{\textbf{Qwen1.5-7B}} \\ \toprule [1pt]
        100\%$^{1}$  & Dense  & 7.95  & 11.93  & 82.45  & 79.05  & 66.14 & 76.90 & 62.25 & 42.83 & 41.60 & 64.46 \\
        \hline
        \multirow{4}{*}{80\% } 
        & ShortGPT \citep{men2024shortgpt}  & 16.88  & 24.88  & 43.98  & 72.69  & 58.41  & 59.11 & 54.50 & 33.70 & 32.20 & 50.66 \\
        & MHA-Drop \citep{he2024matters}   & 14.26  & 22.73  &  59.91 & 75.90  & 58.96  & 67.61 & 61.73 & 41.89 & 37.20 & 57.59 \\
        & FFN-Skip \citep{jaiswal2024ffn} & 94.77  & 123.12  & 45.26  & 59.19  & 51.30  & 36.67 & 36.41 & 22.70 & 28.00 & 39.93 \\
          & \cellcolor{aquamarine}\textbf{\model~} &   \cellcolor{aquamarine}\textbf{18.88} & \cellcolor{aquamarine}\textbf{30.88} & \cellcolor{aquamarine}\textbf{64.50} & \cellcolor{aquamarine}\textbf{73.39} & 
          \cellcolor{aquamarine}\textbf{56.51} & 
          \cellcolor{aquamarine}\textbf{59.98} & \cellcolor{aquamarine}\textbf{56.26} & 
          \cellcolor{aquamarine}\textbf{36.09} &
          \cellcolor{aquamarine}\textbf{38.60} & \cellcolor{aquamarine}\textbf{55.05}\\ 
          \hline 
      \multirow{4}{*}{60\% } 
        & ShortGPT    & 445.24  & 701.1  & 54.55  & 56.08  & 51.07  & 32.49 & 32.37 & 24.23 & 28.40 & 39.89 \\
         & MHA-Drop & 628.12  & 676.62  & 45.87  & 54.45  & 51.45  & 33.16 & 33.08 & 25.67 & 29.59 & 39.05 \\
         & FFN-Skip   & 1889780.25  & 2455505.75  & 46.7  & 51.69  & 49.64  & 26.41 & 25.21 & 25.85 & 28.79 & 36.33 \\
          & \cellcolor{aquamarine}\textbf{\model~} &   \cellcolor{aquamarine}\textbf{54.48} & \cellcolor{aquamarine}\textbf{68.33} & \cellcolor{aquamarine}\textbf{54.76} & \cellcolor{aquamarine}\textbf{61.70} & 
          \cellcolor{aquamarine}\textbf{51.07} & 
          \cellcolor{aquamarine}\textbf{39.72} & \cellcolor{aquamarine}\textbf{44.28} & 
          \cellcolor{aquamarine}\textbf{24.32} &
          \cellcolor{aquamarine}\textbf{29.59} & \cellcolor{aquamarine}\textbf{43.64}\\
        
        \midrule [1pt]
       \rowcolor{gray!10}  \multicolumn{12}{c}{\textbf{Qwen2.5-7B}} \\ \toprule [1pt]
        100\%  & Dense  & 6.85 & 11.36  & 84.61  & 79.71  & 73.00  & 78.95  & 77.40 & 51.01 & 47.40 & 70.30 \\ 
        \hline
        \multirow{4}{*}{80\% } 
        & ShortGPT & 523.53 & 2154.89  & 72.20  & 66.59  & 56.35  & 48.50  & 61.99 & 40.27 & 36.40 & 54.62 \\
        & MHA-Drop  & 115.11  & 184.05  & 42.75  & 71.38  & 57.46  & 55.60 & 52.90 & 39.25 & 40.40 & 51.39 \\
        & FFN-Skip   & 141.24  & 175.33 & 48.69  & 61.26  & 53.51  & 42.08 & 45.16 & 31.14 & 30.00 & 44.55 \\
         & \cellcolor{aquamarine}\textbf{\model~} & 
         \cellcolor{aquamarine}\textbf{13.56} &
         \cellcolor{aquamarine}\textbf{20.33} & 
         \cellcolor{aquamarine}\textbf{70.46} & 
         \cellcolor{aquamarine}\textbf{72.74} & 
         \cellcolor{aquamarine}\textbf{60.62} & 
         \cellcolor{aquamarine}\textbf{63.93} &
         \cellcolor{aquamarine}\textbf{57.87} &
         \cellcolor{aquamarine}\textbf{37.29} & 
         \cellcolor{aquamarine}\textbf{35.19} & 
         \cellcolor{aquamarine}\textbf{56.87}\\ \hline 
      \multirow{4}{*}{60\% } 
        & ShortGPT   & 3460.52  & 4107.47  & 38.59  & 54.03  & 52.80  & 27.61 & 26.56 & 23.63 & 25.40 & 35.52\\
         & MHA-Drop   & 9099.49  & 16067.49  & 48.47  & 54.30  & 50.99  & 28.23 & 29.67 & 27.38 & 32.20 & 38.75\\
         & FFN-Skip   & 1628213.25  & 1434617.50  & 45.78  & 52.12  &  48.93 & 26.83 & 24.54 & 27.3 & 27.6 & 36.16\\
          & \cellcolor{aquamarine}\textbf{\model~} &   
          \cellcolor{aquamarine}\textbf{306.13} & 
          \cellcolor{aquamarine}\textbf{423.79} & 
          \cellcolor{aquamarine}\textbf{47.80} & 
          \cellcolor{aquamarine}\textbf{57.99} & 
          \cellcolor{aquamarine}\textbf{51.07} & 
          \cellcolor{aquamarine}\textbf{33.64} & 
          \cellcolor{aquamarine}\textbf{34.33} & 
          \cellcolor{aquamarine}\textbf{26.54} &
          \cellcolor{aquamarine}\textbf{30.80} & 
          \cellcolor{aquamarine}\textbf{40.31}\\
    \bottomrule [1pt]
    \end{tabular}}
    \renewcommand{\arraystretch}{1} 
    \label{tab:appdenix_result}
\end{table*}


\begin{table*}[!ht]
    \centering
    \resizebox{0.85\linewidth}{!}{\begin{tabular}{c|cccc}
    \toprule
       Schemes  & Llama2-7B 80\% & Llama2-7B 60\% & Llama3-8B 80\% & Llama3-8B 60\%\\
       \midrule
       LLMPruner & 35\%         & 45\%       & 35\%         & 45\% \\
       SliceGPT  & 40\%         & 65\%       & 40\%         & 65\% \\
       ShortGPT  & $\sim$37\% & $\sim$75\%   & $\sim$31\%   & $\sim$75\%\\
       MHA-Drop  & $\sim$26\%   & $\sim$32\% & $\sim$12\%   & $\sim$15\%\\
       FFN-Skip  & $\sim$52\%   & $\sim$64\% & $\sim$65\%   & $\sim$81\%\\
       \model~   & $\sim$24\%   & $\sim$30\% & $\sim$31\%   & $\sim$42\%\\
    \bottomrule    
    \end{tabular}}
    \caption{The pruning ratio of model  weight within the memory budget for different heuristics schemes.}
    \label{tab:remove_ratio}
\end{table*}

\begin{figure*}[!t]
    \centering
    \includegraphics[width=0.95\linewidth]{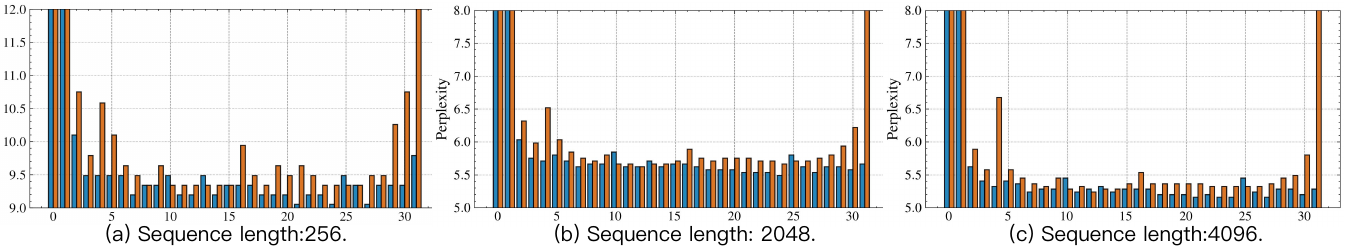}
    \caption{Block sensitivity analysis: removing specific MHA and FFN under diff. sequence length}
    \label{fig:enter-label}
\end{figure*}

\section{Additional Efficiency Analysis}
\label{appendix:efficiency}

\begin{figure*}
    \centering
    \includegraphics[width=0.65\linewidth]{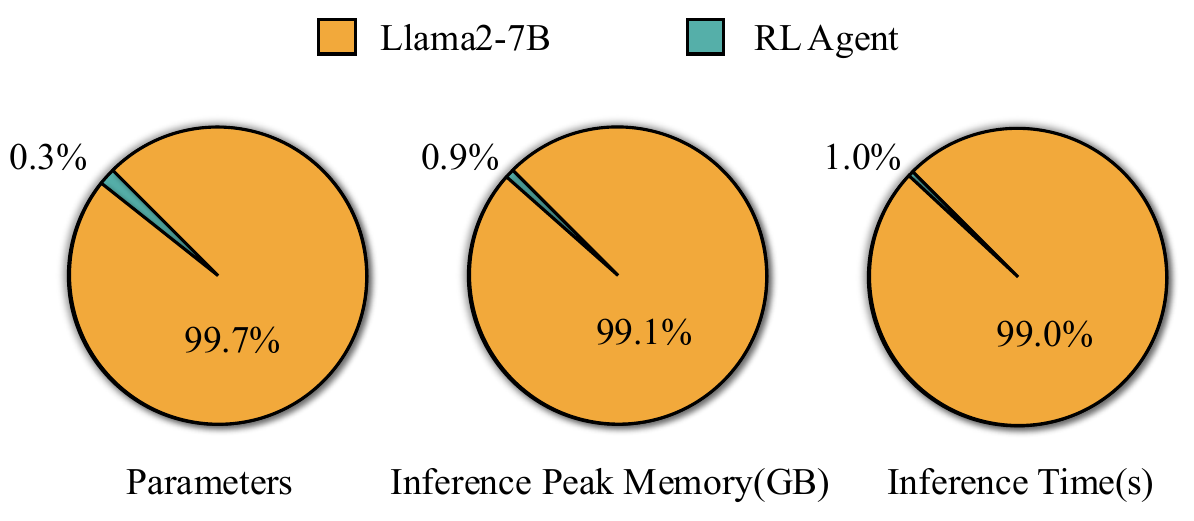}
    \caption{Overhead analysis comparing the RL agent and Llama2-7B in terms of parameter, peak memory usage, and inference latency, illustrating the negligible cost of deploying the RL controller.}
    \label{fig:overhead}
    \vspace{-1em}
\end{figure*}

In this section, we provide a detailed analysis of the computational cost associated with the Greedy Sequential Importance (GSI) metric and compare the end-to-end latency and throughput of our method against baselines on edge devices.

\subsection{Computational Cost of GSI}
\label{appendix:gsi_cost}

For Llama2-7B with 64 blocks, GSI requires evaluating perplexity after each block removal. In our implementation, GSI computes block importance once per model on a large calibration corpus (Alpaca). This process incurs a one-time cost. For instance, calculating GSI scores for Llama2-7B takes approximately 4 hours on a single Nvidia A40 GPU. Crucially, these scores are stored and reused for all subsequent inferences, amortizing the cost to zero during deployment. Users can also opt for a simpler corpus to further reduce this computational overhead. The GSI calculation times for various models are summarized in Table~\ref{table:gsi_time}.

\begin{table}[h]
\centering
\caption{Computational cost of GSI on a single Nvidia A40 GPU.}
\label{table:gsi_time}
\begin{sc}
\begin{tabular}{lcc}
\toprule
\textbf{Model} & \textbf{Total Blocks} & \textbf{GSI Time (h)} \\
\midrule
Llama2-7B   & 64 & 4.0 \\
Llama3-8B   & 64 & 3.9 \\
Qwen1.5-7B  & 64 & 4.5 \\
Qwen2.5-7B  & 56 & 3.5 \\
\bottomrule
\end{tabular}
\end{sc}
\end{table}

\subsection{End-to-End Latency and Throughput Comparison}
\label{appendix:latency&throughput}

Our proposed method is primarily tailored for resource-constrained edge devices rather than high-performance server environments. On such edge platforms, the strict memory limit often serves as the hard constraint that dictates feasibility, whereas servers prioritize aggregate throughput. 

To demonstrate practical efficiency, we conducted additional experiments to evaluate the end-to-end latency and throughput on a representative edge device. The results, presented in Table~\ref{tab:latency&throughput}, demonstrate that our approach achieves lower latency and higher throughput compared to baselines (Dense model, LLM-Pruner~\cite{ma2023llmpruner}, and SliceGPT~\cite{ling2024slimgpt}) by effectively reducing memory consumption and computational overhead.

\begin{table}[h]
\centering
\caption{End-to-end latency and throughput comparison on an edge device.}
\label{tab:latency&throughput}
\begin{sc}
\begin{tabular}{lcc}
\toprule
\textbf{Method} & \textbf{Latency (s)} & \textbf{Throughput (token/s)} \\
\midrule
Dense       & 52.40 & 39.08 \\
LLM-Pruner  & 83.47 & 24.53 \\
SliceGPT    & 64.64 & 31.68 \\
\textbf{RAP (Ours)} & \textbf{43.36} & \textbf{47.23} \\
\bottomrule
\end{tabular}
\end{sc}
\end{table}

\section{Limitation}
\label{appendix:limitation}
Despite its promising results, \model~ still faces several important limitations. First, the Greedy Sequential Importance procedure relies on repeated perplexity measurements over an external corpus (Alpaca), which may become computationally prohibitive for models with tens-of-billions of parameters or for domains lacking a representative calibration set, thereby limiting scalability. Secondly, while the online controller adds negligible inference latency, the offline reinforcement-learning stage still demands several hundred seconds of GPU time and shows sensitivity to the reward coefficients $\alpha$, $\beta$, suggesting non-trivial tuning effort for new hardware or workload profiles. Thirdly, the current state representation tracks only batch size, sequence length and instantaneous memory, omitting latency, energy and heterogeneous device characteristics; as a result, the learned policy may yield sub-optimal trade-offs when such factors dominate deployment objectives. Finally, we note that addressing the challenges of long-context inference, which leads to substantial growth in the KV cache and is often infeasible on resource-constrained devices, is beyond the scope of this paper. Nevertheless, we believe our method's demonstrated efficiency in compressing the KV cache provides a promising foundation for future community efforts in long-context inference compression.

\section{The Use of Large Language Models}
\label{appendix:llm_usage}
We used LLMs solely as a writing-assistance tool to polish our paper (grammar, wording, concision, and minor \LaTeX{} formatting). The LLM did not contribute to research ideation, problem formulation, method design, experiments, data analysis, results, or conclusions, and it was not used to generate citations or technical content. All suggestions were reviewed and, when adopted, edited by the authors, who take full responsibility for the paper’s content; no proprietary data beyond the manuscript text was shared with the tool.




\end{document}